\documentclass[journal]{IEEEtran}
\usepackage{amsmath,amsfonts}
\usepackage{algpseudocode}
\usepackage{algorithmicx,algorithm}

\usepackage{array}
\usepackage[caption=false,font=normalsize,labelfont=sf,textfont=sf]{subfig}
\usepackage{textcomp}
\usepackage{stfloats}
\usepackage{url}
\usepackage{verbatim}
\usepackage{graphicx}
\usepackage{cite}

\usepackage[square,sort&compress,numbers]{natbib}
\usepackage{amsmath}
\usepackage{amsfonts} 
\usepackage{amssymb} 
\usepackage{multirow}
\usepackage{threeparttable}
\usepackage{booktabs}

\graphicspath{{figs/}}

\begin{document}
\title{Continuous Remote Sensing Image Super-Resolution based on Context Interaction in Implicit Function Space}

\author{
Keyan~Chen,~Wenyuan~Li,~Sen~Lei,~Jianqi~Chen,~Xiaolong~Jiang,~Zhengxia~Zou,~and~Zhenwei~Shi$^\star$,~\IEEEmembership{Member,~IEEE}
}

\maketitle

\begin{abstract}

Despite its fruitful applications in remote sensing, image super-resolution is troublesome to train and deploy as it handles different resolution magnifications with separate models. 
Accordingly, we propose a highly-applicable super-resolution framework called FunSR, which settles different magnifications with a unified model by exploiting context interaction within implicit function space.
FunSR composes a functional representor, a functional interactor, and a functional parser.
Specifically, the representor transforms the low-resolution image from Euclidean space to multi-scale pixel-wise function maps; the interactor enables pixel-wise function expression with global dependencies; and the parser, which is parameterized by the interactor's output, converts the discrete coordinates with  additional attributes to RGB values.
Extensive experimental results demonstrate that FunSR reports state-of-the-art performance on both fixed-magnification and continuous-magnification settings, meanwhile, it provides many friendly applications thanks to its unified nature.
\end{abstract}


\IEEEpeerreviewmaketitle

\section{Introduction}

\IEEEPARstart{I}{mage} super-resolution (SR), as a vital technique for elevating image and video spatial resolution, has found vast applications in medical imaging \cite{mahapatra2019image, chen2021super}, surveillance and security \cite{pang2019jcs, zhang2010super}, and remote sensing image analysis \cite{liu2021aerial, merino2007super, lei2017super}, \textit{etc}. Particularly, remote sensing applications including object detection and recognition \cite{ji2019vehicle, courtrai2020small, xie2021super, chen2022degraded}, semantic segmentation \cite{lei2019simultaneous, zhang2021collaborative, chen2021building}, and change detection \cite{chen2021remote} require high-resolution (HR) images with rich high-frequency details and texture information to perform effective image discrimination, analysis, interpretation, and perception. In face of the forbidden cost in image acquisition, SR becomes indispensable for remote sensing image applications.

\begin{figure}[t]
\centering
\resizebox{\linewidth}{!}{
\includegraphics[width=\linewidth]{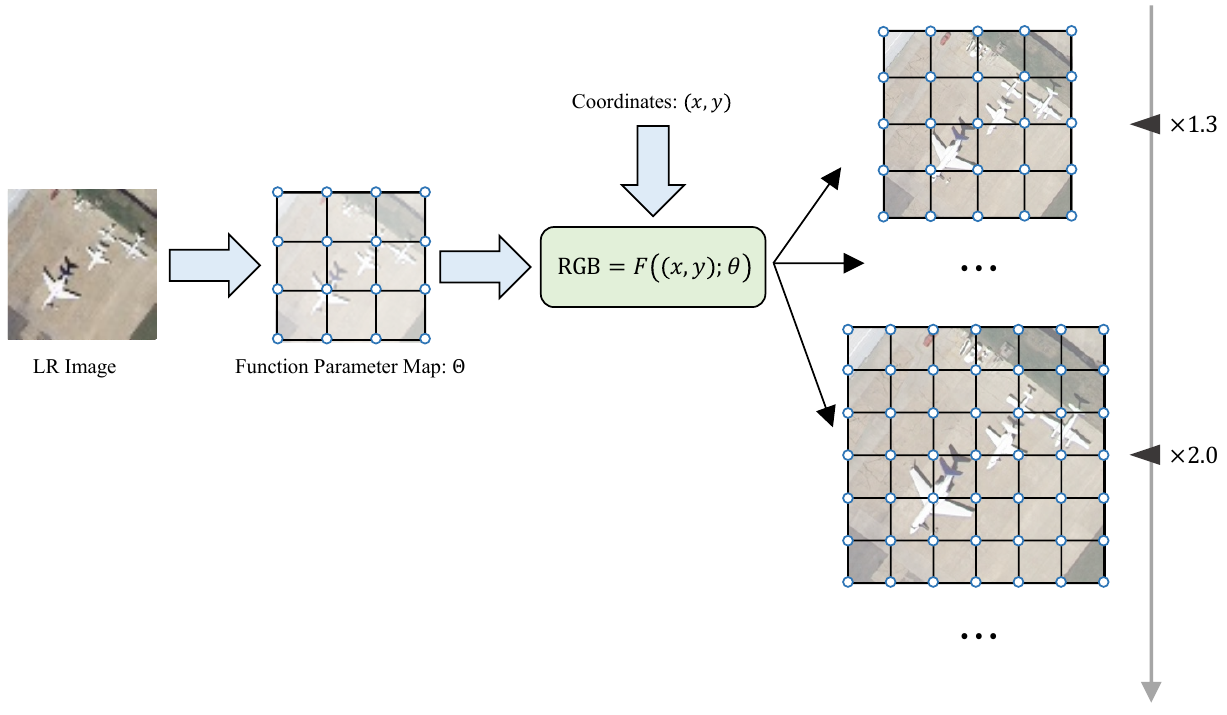}
}
\caption{Our proposed FunSR is capable of producing images of \textit{arbitrary resolution} by \textit{a single trained model}.
FunSR converts location coordinates with some additional attributes, \textit{e.g.}, scale factor, to RGB values using functions parameterized by the transformed LR image.
\label{fig:teaser}}
\end{figure}

Continuous magnification SR is especially vital for remote sensing imagery for three reasons. \textit{First}, imaging real-world surface objects is a process of discretizing continuous signals influenced by the relative acquisition height, so that the objects of the same actual size will show different scales in different images or at different locations. \textit{Second}, multi-source and multi-resolution image processing is the key challenge in remote sensing imagery, which can be partially settled  by continuous magnification.  \textit{Third}, different spatial resolutions might reveal distinct layers and structures of surface objects. It is also critical to efficiently and effectively run the SR network with different magnifications in various application settings.

Albeit favorable applicability, image SR is hard to achieve due to its ill-posed nature, such that one LR image usually corresponds to a large set of HR ones \cite{wang2020deep}. To learn this complicated LR-to-HR mapping, compared with classical learning methods or statistical methods based on edge \cite{xie2015edge, tai2010super}, sparse representation \cite{yang2010image, yang2008image, zhang2015image}, local information encoding \cite{chang2004super, chan2009neighbor}, \textit{etc}., deep learning methods demonstrate greater feature expression ability and better performance. Various deep learning methods have been widely employed and achieved milestone results, ranging from the convolutional neural network (CNN) based methods (\textit{e.g.}, \cite{yamanaka2017fast, tian2020lightweight, tian2020coarse, lei2021hybrid}) to generative adversarial network (GAN) based methods (\textit{e.g.}, \cite{zhang2020supervised, wang2020ultra}) and, more recently, Transformer based methods (\textit{e.g.}, \cite{yang2020learning, lu2022transformer}) and implicit neural representation (INR) based methods (\textit{e.g.}, \cite{chen2021learning, xu2021ultrasr}).

Although these methods have greatly improved the development of image SR, the majority of them adhere to the paradigm of using upsampling operations to achieve the enlargement of LR images, and generally only apply a fixed magnification factor or multiple fixed magnification factors.
According to the position of the upsampling operation, they may be split into the pre-upsampling SR  framework \cite{chang2004super, kim2016accurate} and the post-upsampling one \cite{lim2017enhanced, haut2019remote}.
Most of these methods cannot achieve continuous magnification SR. During the application, they will alter the network structure and retrain the model to meet changing magnification needs, posing significant training, deployment, and storage issues.

Currently, only a few researchers have attempted to use a single model to handle various upscaling factors or continuous factors, and these methods have primarily focused on natural images.
These SR methods can be classified into two folds: conditioned on scale factor and coordinate-based representation.
The previous one frequently designs an upsampling module that is conditioned on the scale factor in order to dynamically rescale the output. It is difficult to get optimal performance on out-of-distribution training magnification and large-scale datasets since they simply use a single number (\textit{i.e.}, the magnification) to regulate the continuous spatial resolution. Furthermore, they can only excel in low-magnification (less than $\times 4$) SR conditions.
INR-based SR breaks the paradigm of explicitly incorporating the magnification factor as an input, which turns the LR image signal into a representation in the coordinate space, then recovers the HR image by querying the pixel-wise features according to the coordinates.
The INR-based SR approach has unique advantages in efficiently modeling continuous differentiable signals and large-scale HR signals.
Most INR-based SR methods are extensions of local implicit image function (LIIF) \cite{chen2021learning}, which simply concatenates the coordinates/periodic transformation of coordinates and local features and predicts the RGB value using a multilayer perceptron (MLP) network.
These methods mainly focus on synthesis from local representation, ignore global semantic coherence, and miss the expression of multi-level features in diverse magnification settings. Furthermore, concatenating coordinates and local features to the same decoding network limits the representation of local high-frequency details and large-scale datasets.

To address the above concerns, we propose FunSR, a method for learning continuous representations of remote sensing images based on context interaction in implicit function space, as illustrated in Fig.~\ref{fig:teaser}.
FunSR is made up of three modules: a functional representer, a functional interactor, and a functional parser, as shown in Fig.~\ref{fig:overall_model}.
The functional representor is designed to convert a low-resolution image from Euclidean space into multi-scale pixel-wise function maps, which can be implemented by feature extraction networks, \textit{e.g.}, ResNet~\cite{simonyan2014very}, EDSR~\cite{lim2017enhanced}, and RDN~\cite{zhang2018residual}.
The functional interactor allows for the expression of pixel-wise functions with global dependencies, \textit{i.e.}, each local function can interact with any other functions at different locations and levels. It contributes to global semantic coherence within the resulted super-resolved HR images.
To fit the image's high-frequency features, the functional parser takes discrete coordinates as input and produces RGB values of the corresponding locations with a periodic activation function \cite{sitzmann2020implicit}. The parameters of the parser are derived from the interactor. We develop a dual-path parser to enhance contextual fusion, \textit{i.e.}, a global parser, and a local parser. The global parser's parameters are shared across locations, whereas the local parser utilizes location-dependent varying parameters.

The main contributions of this paper are as follows:

1) We propose FunSR, a method for learning continuous representations of remote sensing images in implicit function space based on context interaction. FunSR first converts the LR image to a continuous function representation, then takes the HR image's discrete coordinates as input and outputs the RGB values corresponding to the discrete coordinates according to the function. We can achieve arbitrary magnification SR by employing varied sample intervals in the continuous coordinate space.

2)
We propose a functional interactor that allows each pixel-wise function to interact with functions at other locations, therefore enhancing global semantic coherence. Furthermore, we present a dual-path functional parser for generating HR images by parsing coordinates from the global and local levels, improving the feature description ability at different concepts.

3) FunSR is a training, deployment, and storage friendly continuous-scale remote sensing image SR framework. It excels at expressing local high-frequency details, contextual consistency, and large-scale signal generalization. In terms of objective indicators and visual effects, FunSR outperforms other state-of-the-art methods.

The remainder of this paper is structured as follows. Sec. II contains a full description of the related works. Sec. III thoroughly discusses the proposed FunSR. Sec. IV presents quantitative and qualitative results as well as ablation studies. Sec. IV brings this paper to a close.

\section{Related Works}

\subsection{Deep Learning based Remote Sensing Image SR}

Deep learning has made significant progress in the field of remote sensing image SR in recent years \cite{lei2017super, jiang2019edge, lei2021transformer}.
Lei $et~al.$ \cite{lei2017super} propose a local-global-combined network that learns residuals between HR remote sensing images and upscaled LR ones by utilizing the multi-level features of CNN.
Qin $et~al.$ \cite{qin2020remote} design a gradient-aware loss combined with the L1 loss to improve the recovered edges of surface targets.
Wang $et~al.$ \cite{wang2022fenet} construct a lightweight feature enhancement network to achieve a good trade-off between model complexity and performance for remote sensing images.
Chen $et~al.$ \cite{chen2021remote} propose a U-Net like network combined with a split attentional fusion model to obtain HR remote sensing images. 
Moreover, GAN is also introduced to improve the visual super-resolved results for remote sensing LR images. 
Jiang $et~al.$ \cite{jiang2019edge} incorporate the edge-enhancement structure into the traditional GAN framework to weaken the influence caused by noises and artifacts. 
Lei $et~al.$ \cite{lei2019coupled} propose coupled adversarial training with a well-designed discriminator to learn a better discrimination between the super-resolved image and the corresponding ground truth. 
Liu $et~al.$ \cite{liu2021saliency} design a saliency-guided GAN method to improve visual results with additional salient priors.
Some researchers focus on the SR of remote sensing satellite videos. 
Shen $et~al.$ \cite{shen2021deep} combine the multi-frame SR model with an edge-guided single-frame SR for remote sensing video reconstruction. 
These methods are centered on the study of image SR with fixed magnification. In various application scenarios, the network must be redesigned and retrained to accommodate varied magnifications.

\subsection{Image SR with Continuous Magnification}

The SR method of arbitrary magnification has significantly advanced research. It vastly outperforms the previous single-magnification SR method in terms of convenience, by breaking the paradigm of just targeting a single specific integer scale. 
MDSR \cite{lim2017enhanced} presents a multi-scale deep SR network with multiple magnification factors, based on the developed enhanced deep SR network (EDSR) in the paper, but it can only deal with several pre-defined integer magnifications.
Meta-SR \cite{hu2019meta} achieves SR at arbitrary magnification by the designed meta-upscale module, which can utilize the coordinates and scale factors to build the parameters of the convolution kernel. However, using a single simple scale information to condition the entire SR network would restrict the performance.
The architecture of Meta-SR has been further explored and improved in some subsequent works (\textit{e.g.}, ArbRCAN \cite{wang2021learning}, RSI-HFAS \cite{ni2021hierarchical}, RSAN \cite{fu2021residual}). They often design a scale parser module to term the magnification factor as a network conditional input, or an upsampling module to dynamically resize the feature map according to the magnification.
These methods use a single variable (magnification factor) to dynamically modulate the spatial size of the SR image. It is difficult to attain optimal performance in out-of-distribution and large-scale synthesis \cite{chen2021learning}, and these methods still focus on the synthesis of low-upscale factor (less than $\times 4$).

\begin{figure*}[t]
\centering
% \resizebox{宽度}{高度}{对象}
\resizebox{\linewidth}{!}{
\includegraphics[width=\linewidth]{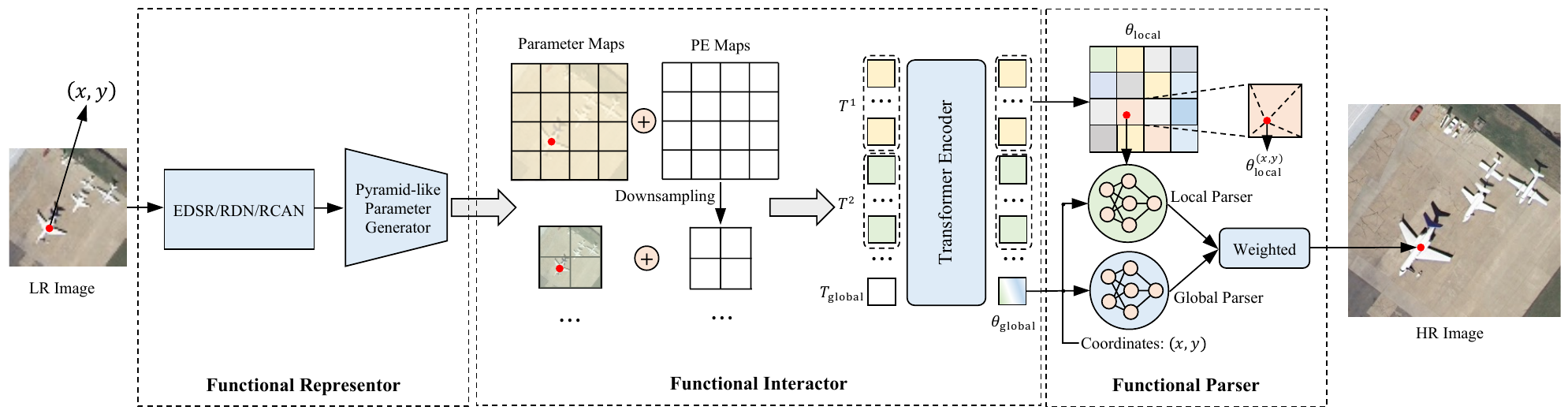}
}
\caption{The outline of the proposed FunSR for continuous magnification remote sensing image SR.
The LR image is first converted to multi-scale parameter maps by the functional representor. Then, we design a functional interactor, \textit{i.e.}, a Transformer encoder, to grasp the effective relationship between functions at different pixel-wise locations and contextual levels. It returns a parameter map with global interaction for the local parser and a semantic parameter vector for the global parser via an additional learnable token. Finally, we weight the RGB value produced by the local and global parsers parameterized with the local parameter map and the global parameter vector, respectively, to generate the final RGB value in the HR image.
}
\label{fig:overall_model}
\end{figure*}

\subsection{Implicit Neural Representations}

INR enables continuous magnification image SR by breaking the paradigm of explicitly employing magnification factors as network input. 
It can recover high-quality SR images of any size by learning continuous image representation in coordinate space and then sampling discrete pixel signals based on different-sized coordinate maps.
INR is essentially a continuously differentiable function that can map properties (\textit{e.g.}, amplitude, intensity, distance) of a space/time point as a function of the related coordinates, \textit{i.e.}, $F_\theta : \mathbb{R}^n \rightarrow \mathbb{R}^m$. Taking representing an image as an example, $F_\theta$, usually an MLP, converts coordinates ($n=2$, the coordinates) to pixel values ($m=3$, RGB values).
It is extensively used to represent objects, images, scenes, \textit{etc}.
Learning implicit neural representations of 2D images lends itself well to image super-resolution challenges, where more detailed (\textit{i.e.}, higher resolution) image representations can be acquired by sampling pixel signals (\textit{i.e.}, RGB values) anywhere in the spatial domain.
Inspired by INR, LIIF \cite{chen2021learning} designs a local implicit image function to achieve continuous image SR, which takes the coordinates and nearby feature representations as input and outputs the RGB value of the corresponding location. Based on LIIF, IPE-LIIF \cite{liu2021enhancing} aggregates local frequency information by positional encoding to improve SR performance. UltraSR \cite{xu2021ultrasr} deeply integrates coordinate encoding with implicit neural representations to improve the accuracy of high-frequency textures.

The aforementioned INR-based SR methods concentrate on constructing an upsampling decoding module by simply concatenating the coordinates/periodic encodings of coordinates and local features together, and predicting the RGB value of the corresponding location using an MLP network.
These methods ignore global semantic consistency and multi-level feature expression at different magnification scenarios. Furthermore, merely concatenating coordinates and local features to the same decoding network will limit the expression of local high-frequency details and large-scale image data while also having an effect on generalization. 
These limitations have a greater impact on SR performance for remote sensing image data, since they have a more complicated distribution, richer details and textures, and a larger data scale than SR datasets frequently utilized in natural images, \textit{e.g.}, face.
To alleviate the aforementioned challenges, we propose FunSR, which transforms the LR image from the image's Euclidean space to the function space, performs multi-scale and global interaction in the function space, and uses local and global functional parsers to get the function value (\textit{i.e.}, RGB value) of the HR image.

\subsection{Transformer for Image Processing}

Transformer was originally developed in natural language processing \cite{vaswani2017attention}. Thanks to its capacity to build long-distance dependencies, researchers have steadily adapted Transformer to computer vision applications in recent years and achieved tremendous success.
ViT \cite{dosovitskiy2020image} verifies that directly applying a pure transformer to sequences of image patches can reach the bar on the performance of a CNN-based model on image classification tasks.
DETR \cite{carion2020end} builds a CNN-Transformer hybrid fully end-to-end detector by combining the benefits of CNN and transformer without anchor generation and non-maximum suppression post-processing.
TTSR \cite{yang2020learning} introduces a novel texture Transformer network for image SR, in which low- and high-resolution images are formulated as queries and keys in a transformer, respectively.
TransENet \cite{lei2021transformer} proposes a transformer-based enhancement network for remote sensing image SR by exploiting features at various levels.
Our proposed FunSR intends to leverage Transformer to enhance the interaction between functions at different spatial locations and different conceptual levels.

\section{Methodology}
In this section, we will introduce the proposed FunSR, a continuous SR method for remote sensing images, including the overview, functional representor, functional interactor, functional parser, and loss function.

\subsection{Overview}
The outline of the proposed FunSR is shown in Fig. \ref{fig:overall_model}. 
Assume we are given a training dataset,
\textit{i.e.}, $\mathcal{D}_{\text{train}} = \{(\mathcal{I}^1_{\text{LR}}, \mathcal{I}^1_{\text{HR}}), \dots, (\mathcal{I}^N_{\text{LR}}, \mathcal{I}^N_{\text{HR}})\}$, where $\mathcal{I}^i_{\text{LR}} \in \mathbb{R}^{H \times W \times 3}$ and $\mathcal{I}^i_{\text{HR}} \in \mathbb{R}^{H^\prime \times W^\prime \times 3}$ refer to the original LR image and its corresponding  ground-truth HR reference. Our goal is to train a model that can process any image from a test set~($\mathcal{I}^k_{\text{LR}} \sim \mathcal{D}_{\text{test}}$), simultaneously obtaining various magnification HR images through coordinate maps of different sizes as follows,
\begin{align}
\begin{split}
    \Theta &= \Phi_{\text{interactor}} \circ \Phi_{\text{representor}} (\mathcal{I}^k_{\text{LR}}) \\
    \mathcal{I}_{\text{HR}} &= \Phi_{\text{parser}} ((x_i, y_i) ; \Theta)
    \label{eq:overall}
\end{split} 
\end{align}
where the image is processed progressively by a functional representor and a functional interactor to generate the local-global interacted parameter map ($\Theta \in \mathbb{R}^{H \times W \times d}$) of the function.
By inputting the discrete coordinates ($(x_i, y_i) \sim \{(x_0, y_0), (x_1, y_1), \cdots\}$) to the functions ($\Phi_{\text{parser}}$ with parameters $\Theta$), we can acquire the RGB value corresponding to the location $(x_i, y_i)$.
Considering that we use different intervals on continuous coordinates to sample discrete coordinate maps of different sizes, we can obtain HR images of different magnifications through a unified function parser.
For simplicity, we will omit the superscript $k$ when describing the proposed model.

\subsection{Functional Representor}
The functional representor is designed to transform LR images of Euclidean space to function space parameters like follows:
\begin{align}
\begin{split}
    \mathcal{M} &= \{\theta^1, \cdots, \theta^l\} =  \Phi_{\text{representor}} (\mathcal{I}_{\text{LR}})
    \\
    &= \Phi_{\text{pyramid}} \circ \Phi_{\text{encoder}}(\mathcal{I}_{\text{LR}})
\end{split}
\end{align}
where $\theta^i$ refers to the parameter map at the $i$-th level. The encoder ($\Phi_{\text{encoder}}$) used in our method is a CNN module without any upscaling layer.
We have adopted some modules in previous works as the encoder directly, including EDSR \cite{lim2017enhanced}, RCAN \cite{zhang2018image}, and RDN \cite{zhang2018residual}. 
In order to allow pixel-level functions to resolve local visual representations at multiple scales and levels, we design a pyramid-like parameter generator ($\Phi_{\text{pyramid}}$). 
For simplicity and efficiency, the structure of $\Phi_{\text{pyramid}}$ follows a downsampling pipeline rather than an upsampling one, based on Cross Stage
Partial Layer (CSPLayer) \cite{wang2020cspnet, bochkovskiy2020yolov4}. 
We use three CSPLayers with a downsampling factor of 2 between each layer, which generates four parameter maps of different sizes, \textit{i.e.},  $\mathcal{M} = \{\theta^i \in \mathbb{R}^{\frac{H}{2^{i-1}} \times \frac{W}{2^{i-1}} \times d} \}, i \in \{ 1,2,3,4 \}$.

\begin{figure}[t]
\centering
\resizebox{\linewidth}{!}{
\includegraphics[width=\linewidth]{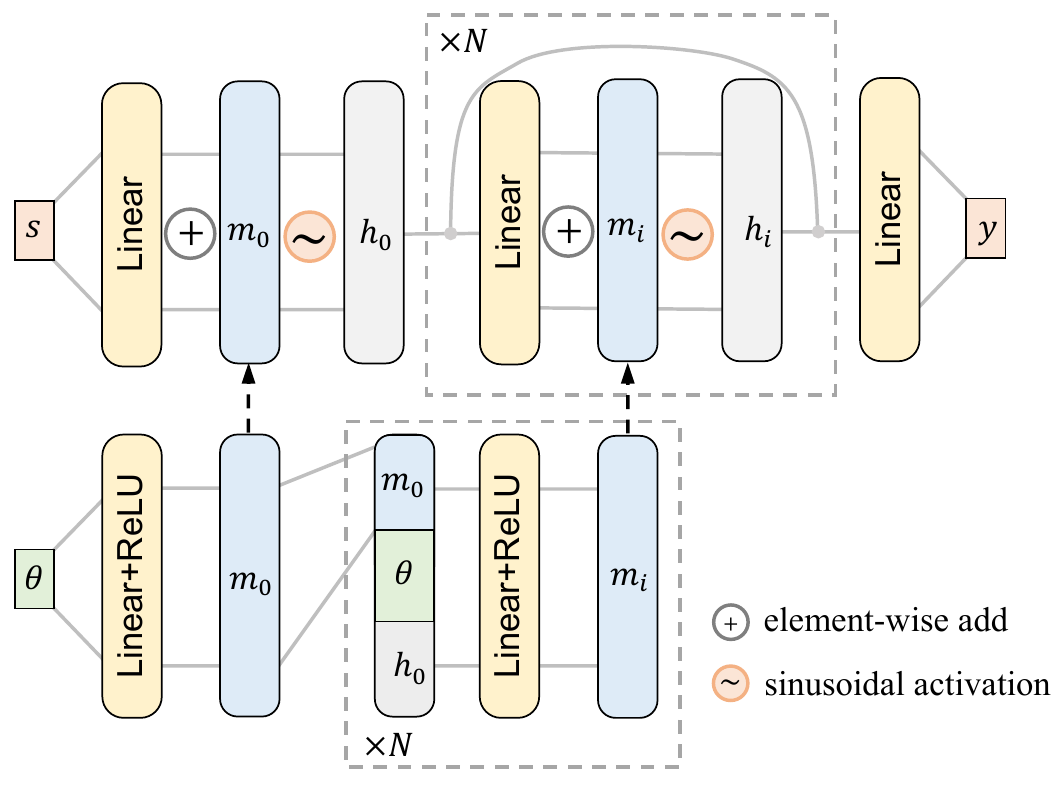}
}
\caption{An illustration of the proposed functional parser, an N-layer MLP with a sinusoidal activation function. 
It takes coordinates with some additional attributes as input ($s$), and returns RGB values ($y$). 
The parser's parameters are modulated by $\theta$.}
\label{fig:parser_model}
\end{figure}

\subsection{Functional Interactor}

We present a functional interactor to enable the expression of local pixel-wise functions to imply global information, \textit{i.e.}, to allow each local pixel-wise function to interact with functions at other locations, even at different conceptual levels. 
The functional interactor contributes to the global semantic coherence of the final parsed HR image. 
Here, we use self-attention based Transformer encoder layers \cite{vaswani2017attention} to implement the interaction between multi-level local functions as follows,
\begin{align}
\begin{split}
    T^i &= \text{Flatten}(\theta^i + \Phi_{\text{samp}}(PE))
    \\
    T &= \text{Cat}([T_{\text{global}}, T^1, \cdots, T^l])
    \\
    \Theta &= \{\theta_{\text{global}}, \theta_{\text{local}}\}
    \\
    &= \Phi_{\text{interactor}}(\mathcal{M})
    = \Phi_{\text{transformer}}(T)
\end{split}
\end{align}
where $PE \in \mathbb{R}^{H \times W \times d}$ is a learnable positional encoding map with the same shape as the first level parameter map, $\theta^1$. $\Phi_{\text{samp}}$ denotes bicubic sampling, making it possible to use the same positional encoding map for other size parameter maps. $\text{Flatten}$ is an operation to flatten the map to vectorized tokens ($T^i$). $T_{\text{global}}$ is a learnable token used to capture global information for subsequent global parsing. $\text{Cat}$ represents vector concatenation to increase the token number. The functional interactor uses N-layer plain transformer encoders ($\Phi_{\text{transformer}}$) to establish dependencies between local functions at different levels, and finally obtains functional parameters that can express global semantic content: global function parameter vector ($\theta_{\text{global}} \in \mathbb{R}^{1 \times d}$) from the first token of the output; local pixel-level function parameter map ($\theta_{\text{local}} \in \mathbb{R}^{H \times W \times d}$) reshaping from the 2-nd to $(H \times W + 1)$-th token of the output.

\subsection{Dual-path Parser} \label{sec:parser}
Through the functional representor ($\Phi_{\text{representor}}$) and the  functional interactor ($\Phi_{\text{interactor}}$), we can represent the LR image as the parameters ($\Theta$) of functions. A naive idea is that these parameters reflect the mapping relationship between image coordinates $(x_i, y_i)$ and image pixel values $\mathcal{I}_{(x_i, y_i)}$, \textit{i.e.}, $\mathcal{I}_{(x_i, y_i)} = f_{\Theta}(x_i,y_i)$. 
We view this as a parsing process from coordinates to pixel values, \textit{i.e.}, term $f_{\Theta}$ as a functional parser ($\Phi_{\text{parser}}$), as illustrated in Eq. \ref{eq:overall}.
Considering that we use different sampling intervals of a fixed range, generally $[-1, 1]$, we can obtain coordinate maps of different sizes. 
Using the functions to parse these coordinates to retrieve the corresponding pixel values, we can generate HR images of any sizes, and achieve continuous magnification SR. Furthermore, the coordinate-based representation method allows us to easily obtain SR of out-of-distribution magnifications.

We present a dual-path coordinate parser that can parse coordinates from both the global and local perspectives, as well as improve image feature descriptions at various levels.
The difference between the two parsers is that: the parameters of the global parser are shared across spatial positions; the parameters of the local parser are dependent on the spatial location. 
The formula description is given below,
\begin{align}
\begin{split}
    \mathcal{I}_{\text{global}} &= \Phi_{\text{global}}((x_i, y_i); \theta_{\text{global}})
    \\
    \theta_{\text{local}}^{(x_i, y_i)} &= \Phi_{\text{interp}}(\theta_{\text{local}}, (x_i, y_i))
    \\
    \mathcal{I}_{\text{local}} &= \Phi_{\text{local}}((x_i, y_i); \theta_{\text{local}}^{(x_i, y_i)})
    \\
    \hat{\mathcal{I}}_{\text{HR}} &= \text{Conv}(\text{Cat}([\mathcal{I}_{\text{global}}, \mathcal{I}_{\text{local}}]))
\end{split}
\end{align}
where $(x_i, y_i)$ is the coordinate, $\Phi_{\text{global}}$ and $\Phi_{\text{local}}$ are the two parsers respectively, and $\Phi_{\text{interp}}$ is an interpolation operation to get the local function parameters at an arbitrary location, \textit{i.e.}, $\theta_{\text{local}}^{(x_i, y_i)} \in \mathbb{R}^{1 \times d}$ from the local parameter map ($\theta_{\text{local}}$).
% Experiments have shown that under the setting of FunSR, various interpolation methods have little impact on performance, so we utilize the nearest interpolation to obtain the local function parameters of a query location for efficiency. 
We utilize the nearest interpolation to obtain the local function parameters of a query location for efficiency.
$\mathcal{I}_{\text{global}}$ and $\mathcal{I}_{\text{local}}$ are the plain HR images parsed from the global and local parsers respectively. 
To produce the required RGB image, we simply concatenate the two to form a 6-channel image and use a convolutional layer with a kernel size of 3 to restore the 3-channel HR image.

Following that, we will go over $\Phi_{\text{global}}$ and $\Phi_{\text{local}}$ in further detail. They are identical in network architecture, so we will introduce the global parser as an example.
When representing pixel values as a function of coordinates, the first challenge lies in the difficulty to use neural networks to fit high-frequency features of images, \textit{e.g.}, edges, and textures.
Fortunately, previous studies have explored the Fourier feature mapping \cite{tancik2020fourier,mildenhall2021nerf} and periodic activation functions \cite{sitzmann2020implicit, chen2022resolution} to learn high-frequency functions in low-dimensional domains. 
We employ a similar MLP architecture with periodic activation functions to form the parser, also known as Sirens \cite{sitzmann2020implicit, chen2022resolution}.
If the parser's parameters are entirely sourced from $\theta_{\text{global}}$, the training of FunSR will suffer from the $\theta_{\text{global}}$ with very high dimension. It's also resource-intensive work.

\begin{figure}[t]
\centering
\resizebox{\linewidth}{!}{
\includegraphics[width=\linewidth]{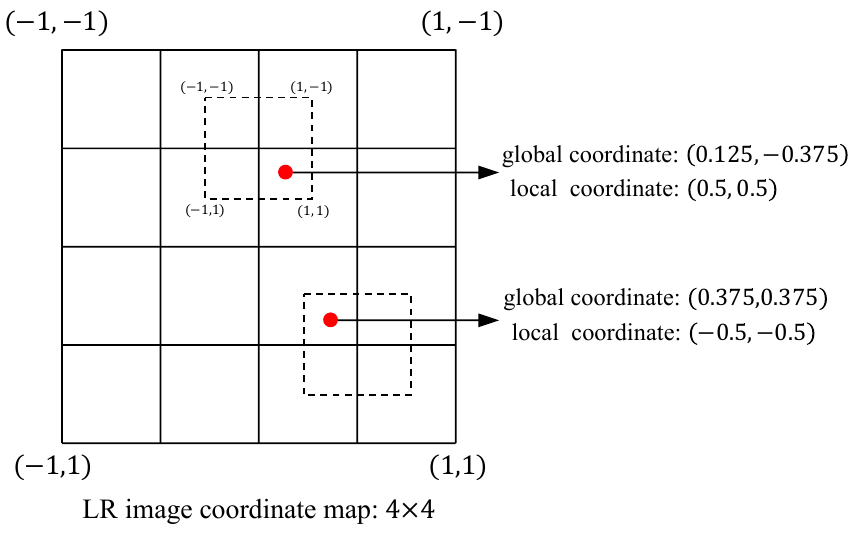}
}
\caption{The global and local coordinates are supposed to be relative to the center of the image and the center of the nearest LR pixel, respectively. All are rescaled to $[-1,1]^2$.}
\label{fig:local_global_coordinates}
\end{figure}

As a result, we only use $\theta_{\text{global}}$ as part of the parser's parameters to regulate the frequency and phase of the periodic activation functions.
In this way, only a small number of parameters (\textit{e.g.}, as few as 64) can be used to perfectly control the whole parsing process, which is friendly for training. 
Our parser performs an N-layer MLP, as illustrated in Fig.~\ref{fig:parser_model}, which can be written recursively as follows,
\begin{align}
\begin{split}
    m_0 &= \text{ReLU}(w_0^m \theta_{\text{global}} + b_0^m)
    \\
    h_0 &= \text{sin}(w_0^h s + m_0)
    \\
    m_i &= \text{ReLU}(w_i^m \text{Cat}([\theta_{\text{global}}, m_{i-1}, h_{i-1}]) + b_i^m)
    \\
    h_i &= \text{sin}(w_i^h h_{i-1} + m_i) + h_{i-1}
    \\
    y & = w_{N}^h h_{N-1} + b_{N}^h
\end{split}
\end{align}
where $w_i^m$, $w_i^h$, $b_i^m$, and $b_i^h$ are the weights and biases, $i$ indicates the $i$-th layer, and $s$ is the input global coordinates $(x_i, y_i)$ with some additional attributes, \textit{e.g.}, scale factor, interpolated RGB value, and local coordinates (here, we refer to the global and local coordinates relative to the center of the image and the center of the nearest LR pixel respectively, as shown in Fig.~\ref{fig:local_global_coordinates}). 
$m_i$ denotes the bias shift to regulate the frequency and phase of the sinusoidal activation function. $h_i$ is the intermediate state of the parser. 
$\text{Cat}$ means vector concatenation. 
The first two formulas describe the modeling process of the first MLP layer, the last one represents the output of the final pixel values of the HR image, and the other two represent the MLP's middle layers.

\subsection{Loss Function}
We train the proposed model using the L1 loss function with $\mathcal{D}_{\text{train}}$ as shown bellow,
\begin{align}
\mathcal{L} = \frac{1}{N\cdot M} \sum_{i=0}^{N} \sum_{j=0}^{M} \parallel \hat{\mathcal{I}}_{\text{HR}}^{(x_i, y_j)} - \mathcal{I}_{\text{HR}}^{(x_i, y_j)} \parallel_{1}
\end{align}
here, we just show the loss function for an image. $\hat{\mathcal{I}}_{\text{HR}}$ and $\mathcal{I}_{\text{HR}}$ are the HR images from our proposed FunSR model and annotations, respectively. $(x_i, y_j)$ is the sampling location for supervised training. $N$ and $M$ denote the width and height of the image.

\section{Experimental Results and Analyses}
\subsection{Experimental Dataset and Settings}
\label{sec:dataset}

In this paper, we use two public remote sensing datasets, UCMecred \cite{ucmerced} and AID \cite{xia2017aid}, to verify the effectiveness of the proposed method. These datasets have been widely used in the field of remote sensing SR \cite{lei2021transformer, haut2019remote, lei2017super}.

\vspace{3pt}
\noindent \textbf{UCMecred Dataset}~\cite{ucmerced}: 
This dataset covers 21 different types of remote sensing scenarios, \textit{e.g.}, agricultural, airplane, baseball diamond, and beach. There are 100 images in each class with a size of $256 \times 256$ pixels, and a spatial resolution of 0.3 m/pixel. We split the dataset into a training set, a validation set, and a test set with a ratio of 6:2:2 for each class.

\vspace{3pt}
\noindent \textbf{AID Dataset}~\cite{xia2017aid}:
This dataset contains 10000 images from 30 different remote sensing scenes, \textit{e.g.}, airport, bareland, church, and dense-residential. All images have an image resolution of $600 \times 600$ pixels and a spatial resolution of 0.5 m/pixel. For the AID dataset, 80\% of the images in each class are selected at random to represent the training set, while the remaining images serve as the test set. Furthermore, we randomly choose 10 images per class in a total of 300 images to form the corresponding validation dataset.

\subsection{Evaluation Protocol and Metrics}
To evaluate the performance of the proposed method, we take two most popular metrics: Peak Signal-to-Noise Ratio (PSNR) and Structural SIMilarity (SSIM).
They are commonly used to objectively assess the quality of image reconstruction \cite{wang2004image, lei2021transformer}. PSNR is defined by Mean Squared Error (MSE), as follows,
\begin{align}
    \text{PSNR} = 10 \times \log_{10} \big (\frac{\text{L}^2}{\text{MSE}} \big)
\end{align}
where $\text{L}$ denotes the achievable largest pixel value, \textit{e.g.}, 255 for an 8-bit image. 

SSIM is more concerned with the perceptual quality of two images, as demonstrated by,
\begin{align}
\begin{split}
    \text{SSIM} &= [l(x,y)]^\alpha \cdot [c(x,y)]^\beta \cdot [s(x,y)]^\gamma \\
    l(x,y) &= \frac{2\mu_x \mu_y + C_1}{\mu_x^2 + \mu_y^2 + C_1} \\
    c(x,y) &= \frac{2\sigma_x \sigma_y + C_2}{\sigma_x^2 + \sigma_y^2 + C_2} \\
    s(x,y) &= \frac{\sigma_{xy} + C_3}{\sigma_x \sigma_y + C_3} \\
\end{split}
\end{align}
where $l(x,y)$, $c(x,y)$ and $s(x,y)$ are measures of luminance, contrast, and structure, respectively, $\mu_x$ and $\mu_y$ are the mean value of $x$ and $y$. $\sigma_x$ and $\sigma_y$ are the variance of $x$ and $y$, $\sigma_{xy}$ is the covariance of $x$ and $y$. $C_1$, $C_2$ and $C_3$ are constants. $\alpha$, $\beta$ and $\gamma$ are usually set to 1.

\subsection{Implementation Details} 

This method focuses on learning continuous representation for remote sensing image SR, \textit{i.e.}, remote sensing image SR with continuous magnifications. In our experiments, we consider the original image to be a real HR reference, and the corresponding LR image is produced via bicubic downsampling.

\subsubsection{Architecture Details}

For the functional representor, we utilize EDSR \cite{lim2017enhanced}, RCAN \cite{zhang2018image}, and RDN \cite{zhang2018residual}, without their upsampling module, as the encoder in our experiments. 
The pyramid-like parameter generator is formed by a 3-layer CSPLayer module with 2, 4, and 6 Darknet Bottleneck blocks in each CSPLayer.
The functional interactor is a 3-layer transformer encoder with 256 input and output dimensions and 512 feedforward dimensions.
As for the parser, the global and local parsers use the same design, but with distinct parameter modulations.
Specifically, the functional parser is made up of a 5-layer MLP (each with 256 hidden units), as described in Sec. \ref{sec:parser}.

\begin{table*}[!htbp] 
\begin{minipage}[b]{\linewidth}
\centering
\caption{Mean PSNR (dB) and SSIM on the UCMerced test dataset with continuous upscale factors.}
\label{tab:uc_sota_allx}
\resizebox{\linewidth}{!}{
\begin{tabular}{c c| *{5}{c} | *{3}{c}}
\toprule
\multicolumn{2}{c|}{\multirow{2}{*}{Method}}
& \multicolumn{5}{c|}{\multirow{1}{*}{In-distribution (PSNR$\uparrow$/SSIM$\uparrow$)}}
& \multicolumn{3}{c}{\multirow{1}{*}{Out-of-distribution (PSNR$\uparrow$/SSIM$\uparrow$)}}
\\
& & $\times$2.0 & $\times$2.5& $\times$3.0& $\times$3.5& $\times$4.0 & $\times$6.0 &  $\times$8.0 & $\times$10.0
\\
\midrule
\multicolumn{2}{c|}{\multirow{1}{*}{Bicubic}} & 31.96/0.9029 & 28.77/0.8356 & 26.68/0.7698 & 25.48/0.7228 & 24.55/0.6803 & 22.04/0.5597 & 20.74/0.4926 & 19.85/0.4527
\\
\multicolumn{2}{c|}{\multirow{1}{*}{SRCNN \cite{dong2015image}}}
&32.71/0.9094 & -
&27.14/0.7824 & -
&25.15/0.6984 & - & - & -
\\
\multicolumn{2}{c|}{\multirow{1}{*}{FSRCNN \cite{dong2016accelerating}}}
&32.04/0.8988 & -
&26.94/0.7756 & -
&24.92/0.6913 & - & - & - 
\\
\multicolumn{2}{c|}{\multirow{1}{*}{LGCNet \cite{lei2017super}}}
&33.48/0.9196 & -
&27.36/0.7913 & -
&25.33/0.7108 & - & - & -
\\
\multicolumn{2}{c|}{\multirow{1}{*}{VDSR \cite{kim2016accurate}}}
&33.97/0.9254 & -
&28.08/0.8126 & -
&25.90/0.7364 & - & - & -
\\
\multicolumn{2}{c|}{\multirow{1}{*}{DCM \cite{haut2019remote}}}
&34.01/0.9263 & -
&28.20/0.8156 & -
&26.02/0.7414 & - & - & -
\\
\multicolumn{2}{c|}{\multirow{1}{*}{TransENet \cite{lei2021transformer}}}
&33.78/0.9233 & -
&28.87/0.8322 & -
&26.98/0.7755 & - & - & -
\\
\midrule
\midrule
\multicolumn{2}{c|}{\multirow{1}{*}{OverNet \cite{behjati2021overnet}}} &33.89/0.9243
&30.43/0.8675
&28.23/0.8142
&27.05/0.7770
&26.05/0.7412
&22.98/0.6017
&21.34/0.5186
&20.24/0.4667
\\
\midrule
\multicolumn{1}{c|}{\multirow{1}{*}{MetaSR \cite{hu2019meta}}} & \multicolumn{1}{c|}{\multirow{6}{*}{EDSR}}
&32.31/0.9066
&29.25/0.8439
&27.16/0.7832
&26.06/0.7404
&25.14/0.7008
&22.62/0.5800
&21.32/0.5122
&20.47/0.4720
\\
\multicolumn{1}{c|}{\multirow{1}{*}{LIIF \cite{chen2021learning}}}& &
33.91/0.9242 
& 30.59/0.8700 
& 28.42/0.8178 
&27.33/0.7862 
& 26.32/0.7487 
& 23.33/0.6204 
&21.82/0.5472 
& 20.79/0.5001
\\
\multicolumn{1}{c|}{\multirow{1}{*}{A-LIIF \cite{li2022adaptive}}}& 
&33.78/0.9224
&30.53/0.8690
&28.36/0.8158
&27.27/0.7841
&26.28/0.7469
&23.32/0.6191
&21.83/0.5469
&20.84/0.5015
\\
\multicolumn{1}{c|}{\multirow{1}{*}{DIINN \cite{nguyen2023single}}}& 
& 34.32/0.9292
&31.00/0.8793
&28.79/0.8281
&27.70/0.7994
&26.64/0.7637
&23.42/0.6259
&21.88/0.5510
&20.82/\textbf{0.5017}
\\
\multicolumn{1}{c|}{\multirow{1}{*}{SADN \cite{wu2021scale}}}& 
&34.18/0.9278
&30.87/0.8772
&28.64/0.8250
&27.53/0.7931
&26.50/0.7575
&23.31/0.6223
&21.73/0.5450
&20.71/0.4985
\\
\multicolumn{1}{c|}{\multirow{1}{*}{FunSR}}&
&\textbf{34.61}/\textbf{0.9318}
&\textbf{31.40}/\textbf{0.8860}
&\textbf{29.19}/\textbf{0.8391}
&\textbf{28.10}/\textbf{0.8095}
&\textbf{27.11}/\textbf{0.7781}
&\textbf{23.62}/\textbf{0.6314}
&\textbf{22.05}/\textbf{0.5531}
&\textbf{20.98}/0.5007
\\
\midrule
\multicolumn{1}{c|}{\multirow{1}{*}{MetaSR \cite{hu2019meta}}} & \multicolumn{1}{c|}{\multirow{6}{*}{RCAN}}
& 33.89/0.9227
&30.59/0.8702
&28.40/0.8166
&27.31/0.7849
&26.33/0.7482
&23.26/0.6162
&21.48/0.5328
&20.36/0.4808
\\
\multicolumn{1}{c|}{\multirow{1}{*}{LIIF \cite{chen2021learning}}}& 
&34.27/0.9282
&31.02/0.8792
&28.92/0.8336
&27.81/0.8008
&26.83/0.7682
&23.54/0.6322
&21.79/0.5516
&20.54/0.4956
\\
\multicolumn{1}{c|}{\multirow{1}{*}{A-LIIF \cite{li2022adaptive}}}&
&34.18/0.9268
&30.94/0.8775
&28.82/0.8305
&27.71/0.7972
&26.72/0.7625
&23.47/0.6302
&21.86/0.5539
&20.77/0.5002
\\
\multicolumn{1}{c|}{\multirow{1}{*}{DIINN \cite{nguyen2023single}}}& 
&34.71/0.9323
&31.47/0.8871
&29.33/0.8430
&28.20/0.8124
&27.16/0.7792
&23.44/0.6300
&21.54/0.5408
&20.38/0.4872
\\
\multicolumn{1}{c|}{\multirow{1}{*}{ArbRCAN \cite{wang2021learning}}}& 
& 34.72/0.9328
&31.39/0.8866
&29.20/0.8393
&28.10/0.8089
&27.10/0.7772
&23.24/0.6116
&21.48/0.5256
&20.42/0.4746
\\
\multicolumn{1}{c|}{\multirow{1}{*}{FunSR}}&
&\textbf{34.86}/\textbf{0.9341}
&\textbf{31.65}/\textbf{0.8902}
&\textbf{29.41}/\textbf{0.8445}
&\textbf{28.27}/\textbf{0.8156}
&\textbf{27.24}/\textbf{0.7799}
&\textbf{23.65}/\textbf{0.6341}
&\textbf{21.94}/\textbf{0.5553}
&\textbf{20.81}/\textbf{0.5010}
\\
\midrule
\multicolumn{1}{c|}{\multirow{1}{*}{MetaSR \cite{hu2019meta}}} & \multicolumn{1}{c|}{\multirow{6}{*}{RDN}}
&34.23/0.9263
&30.98/0.8778
&28.83/0.8295
&27.73/0.7980
&26.76/0.7654
&23.55/0.6309
&21.83/0.5481
&20.66/0.4945
\\
\multicolumn{1}{c|}{\multirow{1}{*}{LIIF \cite{chen2021learning}}}&
&34.30/0.9277
&31.07/0.8801
&28.94/0.8336
&27.84/0.8021
&26.88/0.7676
&23.63/0.6364
&21.95/0.5569
&20.78/\textbf{0.5040}
\\
\multicolumn{1}{c|}{\multirow{1}{*}{A-LIIF \cite{li2022adaptive}}}&
& 34.19/0.9274
&30.96/0.8781
&28.81/0.8288
&27.68/0.7960
&26.72/0.7615
&23.56/0.6337
&21.91/0.5548
&20.79/0.5034
\\
\multicolumn{1}{c|}{\multirow{1}{*}{DIINN \cite{nguyen2023single}}}& 
&34.68/0.9322
&31.38/0.8851
&29.26/0.8405
&28.11/0.8094
&27.06/0.7744
&23.55/0.6361
&21.90/0.5548
&20.78/0.5029
\\
\multicolumn{1}{c|}{\multirow{1}{*}{SADN \cite{wu2021scale}}}&
&34.57/0.9312
&31.31/0.8840
&29.09/0.8338
&28.01/0.8051
&26.98/0.7712
&23.46/0.6329
&21.67/0.5453
&20.65/0.5000
\\
\multicolumn{1}{c|}{\multirow{1}{*}{FunSR}}&
&\textbf{34.82}/\textbf{0.9341}
&\textbf{31.64}/\textbf{0.8898}
&\textbf{29.41}/\textbf{0.8416}
&\textbf{28.31}/\textbf{0.8113}
&\textbf{27.29}/\textbf{0.7798}
&\textbf{23.69}/\textbf{0.6372}
&\textbf{22.04}/\textbf{0.5575}
&\textbf{20.93}/0.5038
\\
\bottomrule
\end{tabular}
}
\end{minipage}

\vspace{10pt}

\begin{minipage}[b]{\linewidth}
\centering
\caption{Mean PSNR (dB) and SSIM on the AID test dataset with continuous upscale factors.}
\label{tab:aid_sota_allx}
\resizebox{\linewidth}{!}{
\begin{tabular}{c c| *{5}{c} | *{3}{c}}
\toprule
\multicolumn{2}{c|}{\multirow{2}{*}{Method}}
& \multicolumn{5}{c|}{\multirow{1}{*}{In-distribution (PSNR$\uparrow$/SSIM$\uparrow$)}}
& \multicolumn{3}{c}{\multirow{1}{*}{Out-of-distribution (PSNR$\uparrow$/SSIM$\uparrow$)}}
\\
& & $\times$2.0 & $\times$2.5& $\times$3.0& $\times$3.5& $\times$4.0 & $\times$6.0 &  $\times$8.0 & $\times$10.0
\\
\midrule
\multicolumn{2}{c|}{\multirow{1}{*}{Bicubic}} & 34.93/0.9169 & 31.76/0.8560 & 29.69/0.7995 & 28.45/0.7595 & 27.48/0.7245 & 24.66/0.6207 & 23.15/0.5656 & 22.15/0.5302
\\
\multicolumn{2}{c|}{\multirow{1}{*}{SRCNN \cite{dong2015image}}}
&35.70/0.9245 & -
&30.31/0.8151 & -
&28.21/0.7465 & - & - & -
\\
\multicolumn{2}{c|}{\multirow{1}{*}{FSRCNN \cite{dong2016accelerating}}}
&35.10/0.9177 & -
&29.93/0.8051 & -
&28.06/0.7416 & - & - & -
\\
\multicolumn{2}{c|}{\multirow{1}{*}{LGCNet \cite{lei2017super}}}
&36.17/0.9304 & -
&30.59/0.8231 & -
&28.45/0.7563 & - & - & -
\\
\multicolumn{2}{c|}{\multirow{1}{*}{VDSR \cite{kim2016accurate}}}
&36.46/0.9341 & -
&31.01/0.8350 & -
&28.93/0.7743 & - & - & -
\\
\multicolumn{2}{c|}{\multirow{1}{*}{DCM \cite{haut2019remote}}}
&36.55/0.9352 & -
&31.19/0.8396 & -
&29.14/0.7804 & - & - & -
\\
\multicolumn{2}{c|}{\multirow{1}{*}{TransENet \cite{lei2021transformer}}}
&36.56/0.9357 & -
&31.43/0.8467 & -
&29.47/0.7937 & - & - & -
\\
\midrule
\midrule
\multicolumn{2}{c|}{\multirow{1}{*}{OverNet \cite{behjati2021overnet}}} & 36.54/0.9354
&33.31/0.8875
&31.32/0.8435
&30.21/0.8127
&29.26/0.7850
&26.21/0.6814
&24.49/0.6102
&23.35/0.5601
\\
\midrule
\multicolumn{1}{c|}{\multirow{1}{*}{MetaSR \cite{hu2019meta}}} & \multicolumn{1}{c|}{\multirow{6}{*}{EDSR}} & 36.35/0.9326
&33.15/0.8841
&31.17/0.8391
&30.06/0.8074
&29.11/0.7790
&26.18/0.6814
&24.54/0.6182
&23.44/0.5730
\\
\multicolumn{1}{c|}{\multirow{1}{*}{LIIF \cite{chen2021learning}}} & & 36.47/0.9346
&33.27/0.8869
&31.29/0.8429
&30.20/0.8122
&29.25/0.7847
&26.33/0.6895
&24.67/0.6274
&\textbf{23.55}/0.5824
\\
\multicolumn{1}{c|}{\multirow{1}{*}{A-LIIF \cite{li2022adaptive}}}& & 36.43/0.9340
&33.23/0.8860
&31.24/0.8415
&30.14/0.8103
&29.19/0.7825
&26.29/0.6875
&24.64/0.6262
&23.53/0.5818
\\
\multicolumn{1}{c|}{\multirow{1}{*}{DIINN \cite{nguyen2023single}}}& & 36.65/0.9366
&33.40/0.8895
&31.42/0.8461
&30.33/0.8160
&29.37/0.7891
&26.36/0.6933
&24.63/0.6290
&23.47/\textbf{0.5825}
\\
\multicolumn{1}{c|}{\multirow{1}{*}{SADN \cite{wu2021scale}}}& &36.61/0.9363
&33.39/0.8892
&31.40/0.8455
&30.30/0.8153
&29.36/0.7885
&26.36/0.6918
&24.64/0.6277
&23.52/0.5823
\\
\multicolumn{1}{c|}{\multirow{1}{*}{FunSR}}& &\textbf{36.74}/\textbf{0.9382}
&\textbf{33.49}/\textbf{0.8925}
&\textbf{31.53}/\textbf{0.8499}
&\textbf{30.44}/\textbf{0.8205}
&\textbf{29.51}/\textbf{0.7948}
&\textbf{26.46}/\textbf{0.6969}
&\textbf{24.72}/\textbf{0.6298}
&23.53/0.5816
\\
\midrule
\multicolumn{1}{c|}{\multirow{1}{*}{MetaSR \cite{hu2019meta}}} & \multicolumn{1}{c|}{\multirow{6}{*}{RCAN}} & 36.75/0.9373
&33.51/0.8918
&31.54/0.8496
&30.46/0.8202
&29.54/0.7942
&26.59/0.7008
&24.87/0.6359
&23.70/0.5874
\\
\multicolumn{1}{c|}{\multirow{1}{*}{LIIF \cite{chen2021learning}}}& & 36.80/0.9383
&33.57/0.8930
&31.60/0.8511
&30.53/0.8219
&29.61/0.7962
&26.68/0.7046
&24.98/0.6419
&\textbf{23.81}/0.5942
\\
\multicolumn{1}{c|}{\multirow{1}{*}{A-LIIF \cite{li2022adaptive}}}&
&36.78/0.9382
&33.55/0.8926
&31.58/0.8505
&30.52/0.8216
&29.59/0.7960
&26.66/0.7039
&24.96/0.6415
&23.79/\textbf{0.5943}
\\
\multicolumn{1}{c|}{\multirow{1}{*}{DIINN \cite{nguyen2023single}}}& &36.97/0.9400
&33.70/0.8952
&31.71/0.8535
&30.63/0.8247
&29.70/0.7993
&26.68/0.7059
&24.91/0.6408
&23.70/0.5920
\\
\multicolumn{1}{c|}{\multirow{1}{*}{ArbRCAN \cite{wang2021learning}}}& 
& 37.11/0.9410
&33.83/0.8969
&31.78/0.8548
&30.66/0.8255
&29.72/0.8001
&26.44/0.6943
&24.57/0.6179
&23.38/0.5652
\\
\multicolumn{1}{c|}{\multirow{1}{*}{FunSR}}&
&\textbf{37.16}/\textbf{0.9417}
&\textbf{33.86}/\textbf{0.8982}
&\textbf{31.82}/\textbf{0.8554}
&\textbf{30.73}/\textbf{0.8272}
&\textbf{29.83}/\textbf{0.8017}
&\textbf{26.75}/\textbf{0.7063}
&\textbf{25.03}/\textbf{0.6427}
&23.79/0.5938
\\
\midrule
\multicolumn{1}{c|}{\multirow{1}{*}{MetaSR \cite{hu2019meta}}} & \multicolumn{1}{c|}{\multirow{6}{*}{RDN}}
& 36.77/0.9374
&33.52/0.8917
&31.55/0.8497
&30.48/0.8204
&29.56/0.7943
&26.62/0.7017
&24.90/0.6373
&23.72/0.5886
\\
\multicolumn{1}{c|}{\multirow{1}{*}{LIIF \cite{chen2021learning}}}&
& 36.81/0.9382
&33.57/0.8927
&31.60/0.8505
&30.52/0.8215
&29.60/0.7957
&26.68/0.7039
&24.98/0.6414
&\textbf{23.80}/\textbf{0.5939}
\\
\multicolumn{1}{c|}{\multirow{1}{*}{A-LIIF \cite{li2022adaptive}}}&
& 36.76/0.9379
&33.51/0.8917
&31.55/0.8496
&30.48/0.8203
&29.55/0.7944
&26.63/0.7022
&24.93/0.6397
&23.76/0.5927
\\
\multicolumn{1}{c|}{\multirow{1}{*}{DIINN \cite{nguyen2023single}}}&
&36.98/0.9398
&33.71/0.8950
&31.71/0.8533
&30.62/0.8243
&29.68/0.7987
&26.65/0.7042
&24.88/0.6387
&23.68/0.5899
\\
\multicolumn{1}{c|}{\multirow{1}{*}{SADN \cite{wu2021scale}}}& 
& 36.84/0.9386
&33.58/0.8930
&31.60/0.8507
&30.53/0.8217
&29.60/0.7961
&26.56/0.7016
&24.81/0.6367
&23.63/0.5890
\\
\multicolumn{1}{c|}{\multirow{1}{*}{FunSR}}&
&\textbf{37.01}/\textbf{0.9406}
&\textbf{33.79}/\textbf{0.8968}
&\textbf{31.81}/\textbf{0.8557}
&\textbf{30.73}/\textbf{0.8269}
&\textbf{29.82}/\textbf{0.8018}
&\textbf{26.70}/\textbf{0.7043}
&\textbf{25.01}/\textbf{0.6417}
&23.77/0.5923
\\
\bottomrule
\end{tabular}
}
\end{minipage}
\end{table*}

\begin{table*}[!htpb] 
\centering
\caption{Mean PSNR (dB) and SSIM of each scene class on the UCMerced test dataset with the upscale factor of $\times 4$.}
\label{tab:uc_sota_classes}
\resizebox{0.92\linewidth}{!}{
\begin{tabular}{c| *{8}{c}}
\toprule
class & Bicubic & SRCNN \cite{dong2015image} & FSRCNN \cite{dong2016accelerating} & LGCNet \cite{lei2017super} & VDSR \cite{kim2016accurate} &DCM \cite{haut2019remote} & TransENet \cite{lei2021transformer}
\\
\midrule
1 & 22.95/0.3882 & 23.55/0.4065 & 23.63/0.4239 & 24.06/0.4868 & 25.26/0.5739 & 26.17/0.6465 & \textbf{26.91}/\textbf{0.6757}
\\
2 & 24.07/0.7093 & 24.62/0.7324 & 24.31/0.7208 & 24.77/0.7403 & 25.26/0.7575 & 25.24/0.7563 & 26.08/0.7825
\\
3 & 30.79/0.7901 & 31.69/0.8043 & 31.34/0.7988 & 31.81/0.8087 & 32.08/0.8166 & 32.11/0.8165 & 32.57/0.8271
\\
4 & 30.98/0.8101 & 31.43/0.8125 & 31.23/0.8103 & 31.48/0.8160 & 31.58/0.8195 & 31.63/0.8203 & 32.28/\textbf{0.8449}
\\
5 & 21.73/0.6908 & 22.28/0.7247 & 22.11/0.7160 & 22.39/0.7350 & 22.82/0.7543 & 22.83/0.7525 & 23.33/0.7743
\\
6 & 23.29/0.6629 & 23.43/0.6576 & 23.38/0.6549 & 23.56/0.6688 & 23.78/0.6876 & 23.74/0.6838 & 24.30/\textbf{0.7189}
\\
7 & 22.08/0.6731 & 22.74/0.7077 & 22.48/0.6937 & 22.92/0.7200 & 23.61/0.7472 & 23.55/0.7463 & 24.97/0.7964
\\
8 & 26.28/0.6255 & 26.61/0.6225 & 26.54/0.6200 & 26.68/0.6283 & 26.99/0.6486 & 26.98/0.6492 & 27.41/0.6766
\\
9 & 24.51/0.7085 & 25.40/0.7403 & 24.94/0.7225 & 25.74/0.7545 & 27.32/0.8000 & 27.68/0.8110 & 29.11/0.8502
\\
10 & 29.43/0.7579 & 30.26/0.7737 & 29.80/0.7649 & 30.43/0.7794 & 30.84/0.7904 & 30.86/0.7924 & 31.53/\textbf{0.8096}
\\
11 & 19.80/0.6932 & 20.53/0.7295 & 20.31/0.7198 & 20.70/0.7422 & 21.27/0.7684 & 21.52/0.7744 & 22.75/0.8181
\\
12 & 24.31/0.7020 & 24.78/0.7138 & 24.61/0.7072 & 24.91/0.7205 & 25.34/0.7394 & 25.34/0.7408 & 26.27/0.7760
\\
13 & 21.74/0.6102 & 22.21/0.6289 & 22.10/0.6235 & 22.33/0.6383 & 22.75/0.6592 & 22.69/0.6564 & 23.61/0.6941
\\
14 & 22.57/0.7198 & 23.07/0.7420 & 22.82/0.7288 & 23.28/0.7516 & 23.99/0.7796 & 24.02/0.7776 & 25.31/0.8206
\\
15 & 22.80/0.6744 & 23.64/0.7103 & 23.38/0.6960 & 23.94/0.7246 & 25.15/0.7737 & 25.53/0.7866 & 27.20/0.8328
\\
16 & 19.06/0.6417 & 19.16/0.6505 & 19.06/0.6446 & 19.27/0.6630 & 19.82/0.6962 & 19.88/0.6931 & 21.72/0.7778
\\
17 & 26.55/0.6757 & 27.34/0.6856 & 27.24/0.6840 & 27.46/0.6907 & 27.85/0.7051 & 27.82/0.7067 & 28.18/0.7248
\\
18 & 26.98/0.7105 & 28.00/0.7321 & 27.45/0.7212 & 28.36/0.7415 & 28.96/0.7608 & 29.14/0.7625 & 30.72/0.7945
\\
19 & 24.13/0.6050 & 24.51/0.6160 & 24.31/0.6110 & 24.62/0.6225 & 24.91/0.6375 & 24.93/0.6354 & 25.33/0.6532
\\
20 & 25.00/0.7395 & 25.79/0.7667 & 25.52/0.7571 & 25.96/0.7763 & 26.64/0.7977 & 26.78/0.7995 & 27.99/0.8355
\\
21 & 26.58/0.6973 & 27.04/0.7078 & 26.79/0.6993 & 27.16/0.7173 & 27.79/0.7513 & 27.95/0.7614 & 28.91/0.8014
\\
\midrule
Average & 24.55/0.6803 & 25.15/0.6984 & 24.92/0.6913 & 25.33/0.7108 & 25.90/0.7364 & 26.02/0.7414 & 26.98/0.7755
\\
\midrule
\midrule
class & OverNet \cite{behjati2021overnet} & MetaSR \cite{hu2019meta} & LIIF \cite{chen2021learning} & A-LIIF \cite{li2022adaptive} & DIINN \cite{nguyen2023single} & SADN \cite{wu2021scale} & FunSR
\\
\midrule
1 & 25.27/0.5752 & 26.01/0.6230 & 25.78/0.5886 & 25.25/0.5441 & 25.03/0.5609 & 24.76/0.5375 & 25.22/0.5659
\\
2 & 25.38/0.7618 & 25.96/0.7763 & 26.09/0.7805 & 26.00/0.7774 & 26.32/0.7868 & 26.31/0.7851 & \textbf{26.43}/\textbf{0.7892}
\\
3 & 32.19/0.8186 & 32.57/0.8253 & 32.60/0.8257 & 32.55/0.8247 & 32.73/0.8286 & 32.66/0.8279 & \textbf{32.74}/\textbf{0.8299}
\\
4 & 31.66/0.8225 & 31.94/0.8299 & 32.04/0.8333 & 31.94/0.8304 & 32.15/0.8397 & 32.12/0.8380 & \textbf{32.35}/0.8445
\\
5 & 23.06/0.7631 & 23.53/0.7794 & 23.59/0.7821 & 23.47/0.7787 & 23.55/0.7846 & 23.58/0.7846 & \textbf{23.87}/\textbf{0.7919}
\\
6 & 23.75/0.6837 & 24.05/0.7014 & 24.13/0.7052 & 24.07/0.7018 & 24.18/0.7118 & 24.17/0.7103 & \textbf{24.33}/0.7186
\\
7 & 23.78/0.7572 & 24.61/0.7834 & 24.85/0.7913 & 24.62/0.7845 & 25.09/0.8000 & 25.00/0.7972 & \textbf{25.39}/\textbf{0.8069}
\\
8 & 26.90/0.6446 & 27.30/0.6657 & 27.33/0.6670 & 27.30/0.6651 & 27.38/0.6730 & 27.36/0.6722 & \textbf{27.47}/\textbf{0.6767}
\\
9 & 27.83/0.8169 & 28.74/0.8369 & 28.98/0.8419 & 28.63/0.8372 & 29.36/0.8542 & 29.17/0.8478 & \textbf{29.66}/\textbf{0.8579}
\\
10 & 30.83/0.7914 & 31.22/0.8011 & 31.29/0.8021 & 31.21/0.8008 & 31.43/0.8064 & 31.40/0.8060 & \textbf{31.57}/0.8095
\\
11 & 21.41/0.7754 & 22.57/0.8096 & 22.81/0.8166 & 22.61/0.8115 & 23.12/0.8261 & 23.01/0.8239 & \textbf{23.45}/\textbf{0.8319}
\\
12 & 25.55/0.7479 & 26.18/0.7723 & 26.35/0.7793 & 26.17/0.7724 & 26.53/0.7888 & 26.52/0.7857 & \textbf{26.87}/\textbf{0.7966}
\\
13 & 22.80/0.6631 & 23.42/0.6853 & 23.49/0.6879 & 23.39/0.6835 & 23.68/0.6981 & 23.62/0.6959 & \textbf{23.89}/\textbf{0.7047}
\\
14 & 24.23/0.7875 & 25.19/0.8153 & 25.36/0.8212 & 25.21/0.8172 & 25.59/0.8298 & 25.51/0.8273 & \textbf{25.78}/\textbf{0.8336}
\\
15 & 25.71/0.7928 & 27.12/0.8311 & 27.38/0.8361 & 27.11/0.8299 & 27.92/0.8476 & 27.66/0.8430 & \textbf{28.21}/\textbf{0.8527}
\\
16 & 19.72/0.6912 & 21.10/0.7536 & 21.38/0.7657 & 21.06/0.7539 & 21.92/0.7870 & 21.82/0.7837 & \textbf{22.49}/\textbf{0.8048}
\\
17 & 27.90/0.7074 & 28.21/0.7186 & 28.22/0.7187 & 28.22/0.7174 & \textbf{28.33}/0.7233 & 28.31/0.7234 & 28.32/\textbf{0.7269}
\\
18 & 29.18/0.7659 & 30.13/0.7837 & 30.33/0.7881 & 30.13/0.7829 & 30.72/0.7963 & 30.75/0.7948 & \textbf{31.25}/\textbf{0.8009}
\\
19 & 25.04/0.6382 & 25.38/0.6504 & 25.43/0.6512 & 25.35/0.6492 & 25.48/0.6573 & 25.40/0.6547 & \textbf{25.57}/\textbf{0.6597}
\\
20 & 26.98/0.8064 & 27.94/0.8302 & 28.05/0.8346 & 27.93/0.8304 & 28.43/0.8464 & 28.29/0.8429 & \textbf{28.66}/\textbf{0.8508}
\\
21 & 27.91/0.7547 & 28.90/0.7999 & 28.98/0.8034 & 28.89/0.7984 & 29.32/0.8165 & 29.22/0.8131 & \textbf{29.55}/\textbf{0.8224}
\\
\midrule
Average & 26.05/0.7412 & 26.76/0.7654 & 26.88/0.7676 & 26.72/0.7615 & 27.06/0.7744 & 26.98/0.7712 & \textbf{27.29}/\textbf{0.7798}
\\
\bottomrule
\end{tabular}
}
\end{table*}

\begin{table*}[!htpb]
\centering
\caption{Mean PSNR (dB) and SSIM of each scene class on the AID test dataset with the upscale factor of $\times 4$.}
\label{tab:aid_sota_classes}
\resizebox{0.92\linewidth}{!}{
\begin{tabular}{c| *{8}{c}}
\toprule
class & Bicubic & SRCNN \cite{dong2015image} & FSRCNN \cite{dong2016accelerating} & LGCNet \cite{lei2017super} & VDSR \cite{kim2016accurate} &DCM \cite{haut2019remote} & TransENet \cite{lei2021transformer}
\\
\midrule
1 & 26.20/0.7252 & 26.96/0.7469 & 26.90/0.7433 & 27.21/0.7564 & 27.65/0.7740 & 27.87/0.7815 & 28.19/0.7942
\\
2 & 34.91/0.8320 & 34.95/0.8327 & 34.70/0.8299 & 35.01/0.8342 & 35.26/0.8395 & 35.33/0.8402 & 35.18/0.8375
\\
3 & 33.06/0.8443 & 33.92/0.8584 & 33.50/0.8526 & 34.20/0.8641 & 34.69/0.8728 & 34.82/0.8752 & 35.05/0.8787
\\
4 & 31.42/0.7812 & 31.71/0.7886 & 31.53/0.7856 & 31.80/0.7917 & 32.11/0.7982 & 32.24/0.8004 & 32.26/0.8035
\\
5 & 27.43/0.7576 & 28.58/0.7874 & 28.40/0.7827 & 29.20/0.8035 & 30.35/0.8358 & 30.67/0.8454 & 31.49/0.8624
\\
6 & 26.16/0.7301 & 27.17/0.7626 & 26.99/0.7564 & 27.60/0.7767 & 28.47/0.7993 & 28.79/0.8092 & 29.54/0.8302
\\
7 & 20.87/0.5550 & 21.91/0.6053 & 21.81/0.5972 & 22.30/0.6281 & 22.88/0.6580 & 23.15/0.6695 & 23.48/0.6957
\\
8 & 24.67/0.7020 & 25.25/0.7216 & 25.21/0.7173 & 25.38/0.7284 & 25.73/0.7456 & 25.91/0.7514 & 26.13/0.7650
\\
9 & 21.74/0.5905 & 22.32/0.6166 & 22.26/0.6105 & 22.45/0.6279 & 22.77/0.6496 & 22.94/0.6563 & 23.22/0.6827
\\
10 & 36.54/0.8724 & 37.07/0.8792 & 36.79/0.8763 & 37.21/0.8821 & 37.28/0.8837 & 37.41/0.8866 & 37.32/0.8855
\\
11 & 32.42/0.8119 & 33.33/0.8296 & 33.17/0.8264 & 33.68/0.8376 & 34.28/0.8518 & 34.58/0.8594 & 34.71/0.8633
\\
12 & 26.77/0.6171 & 27.10/0.6154 & 27.04/0.6116 & 27.19/0.6199 & 27.39/0.6330 & 27.47/0.6348 & 27.46/0.6507
\\
13 & 23.75/0.6569 & 24.66/0.6915 & 24.50/0.6840 & 25.02/0.7077 & 25.69/0.7379 & 26.00/0.7513 & 26.48/0.7790
\\
14 & 32.40/0.7263 & 32.88/0.7386 & 32.78/0.7377 & 32.93/0.7423 & 33.09/0.7470 & 33.21/0.7483 & 33.15/0.7523
\\
15 & 23.73/0.6017 & 24.47/0.6383 & 24.35/0.6331 & 24.70/0.6542 & 25.20/0.6789 & 25.37/0.6846 & 25.67/0.7043
\\
16 & 27.22/0.6972 & 27.75/0.7065 & 27.68/0.7030 & 27.80/0.7097 & 27.89/0.7175 & 27.96/0.7187 & 27.89/0.7268
\\
17 & 27.68/0.7320 & 28.32/0.7457 & 28.24/0.7411 & 28.42/0.7507 & 28.65/0.7604 & 28.68/0.7609 & 28.71/0.7695
\\
18 & 21.30/0.6982 & 21.74/0.7250 & 21.61/0.7174 & 21.90/0.7363 & 22.56/0.7648 & 22.97/0.7797 & 24.36/0.8247
\\
19 & 34.50/0.8506 & 35.27/0.8635 & 34.99/0.8588 & 35.56/0.8701 & 36.30/0.8824 & 36.55/0.8869 & 37.16/0.8953
\\
20 & 36.30/0.8946 & 37.32/0.9029 & 36.93/0.9011 & 37.56/0.9062 & 37.87/0.9109 & 37.98/0.9125 & 38.01/0.9109
\\
21 & 25.07/0.7859 & 25.83/0.8088 & 25.72/0.8038 & 26.07/0.8178 & 26.52/0.8322 & 26.80/0.8393 & 27.28/0.8546
\\
22 & 24.72/0.6747 & 25.65/0.7152 & 25.54/0.7097 & 26.11/0.7374 & 27.04/0.7817 & 27.43/0.7998 & 28.12/0.8271
\\
23 & 24.86/0.6935 & 25.33/0.7120 & 25.21/0.7053 & 25.49/0.7210 & 25.97/0.7399 & 26.13/0.7440 & 26.38/0.7605
\\
24 & 28.96/0.7334 & 29.65/0.7454 & 29.54/0.7422 & 29.80/0.7507 & 30.00/0.7595 & 30.05/0.7599 & 30.06/0.7647
\\
25 & 23.40/0.6653 & 24.09/0.6924 & 23.98/0.6838 & 24.29/0.7031 & 24.71/0.7237 & 24.91/0.7288 & 25.31/0.7504
\\
26 & 22.68/0.5717 & 23.47/0.6086 & 23.41/0.6036 & 23.73/0.6229 & 24.11/0.6411 & 24.27/0.6442 & 24.43/0.6596
\\
27 & 27.60/0.7451 & 28.19/0.7747 & 28.00/0.7671 & 28.45/0.7863 & 29.07/0.8049 & 29.22/0.8091 & 29.60/0.8254
\\
28 & 30.47/0.8352 & 32.08/0.8570 & 31.83/0.8520 & 32.33/0.8618 & 32.95/0.8707 & 33.22/0.8762 & 34.02/0.8892
\\
29 & 22.44/0.6372 & 23.40/0.6784 & 23.29/0.6713 & 23.72/0.6947 & 24.23/0.7199 & 24.52/0.7292 & 25.04/0.7514
\\
30 & 24.96/0.6497 & 25.88/0.6865 & 25.76/0.6820 & 26.33/0.7091 & 27.18/0.7537 & 27.48/0.7684 & 28.11/0.7986
\\
\midrule
Average & 27.48/0.7245 & 28.21/0.7465 & 28.06/0.7416 & 28.45/0.7563 & 28.93/0.7743 & 29.14/0.7804 & 29.47/0.7937
\\
\midrule
\midrule
class & OverNet \cite{behjati2021overnet} & MetaSR \cite{hu2019meta} & LIIF \cite{chen2021learning} & A-LIIF \cite{li2022adaptive} & DIINN \cite{nguyen2023single} & SADN \cite{wu2021scale} & FunSR
\\
\midrule
1 & 27.96/0.7853 & 28.24/0.7950 & 28.31/0.7961 & 28.25/0.7947 & 28.42/0.7987 & 28.32/0.7960 & \textbf{28.62}/\textbf{0.8037}
\\
2 & 35.39/0.8411 & 35.43/0.8424 & 35.43/0.8429 & 35.41/0.8424 & 35.45/0.8434 & 35.43/0.8426 & \textbf{35.51}/\textbf{0.8463}
\\
3 & 34.87/0.8769 & 35.12/0.8815 & 35.15/0.8818 & 35.11/0.8812 & 35.22/0.8831 & 35.18/0.8825 & \textbf{35.34}/\textbf{0.8872}
\\
4 & 32.33/0.8027 & 32.43/0.8048 & 32.43/0.8054 & 32.42/0.8051 & 32.47/0.8065 & 32.43/0.8056 & \textbf{32.56}/\textbf{0.8109}
\\
5 & 30.93/0.8502 & 31.46/0.8619 & 31.49/0.8627 & 31.44/0.8615 & 31.63/0.8658 & 31.49/0.8629 & \textbf{31.79}/\textbf{0.8692}
\\
6 & 28.98/0.8123 & 29.59/0.8283 & 29.64/0.8314 & 29.56/0.8281 & 29.78/0.8354 & 29.63/0.8303 & \textbf{30.04}/\textbf{0.8427}
\\
7 & 23.31/0.6780 & 23.60/0.6920 & 23.62/0.6950 & 23.60/0.6923 & 23.74/0.7005 & 23.66/0.6966 & \textbf{23.88}/\textbf{0.7077}
\\
8 & 25.97/0.7552 & 26.24/0.7662 & 26.26/0.7672 & 26.21/0.7653 & 26.32/0.7695 & 26.26/0.7672 & \textbf{26.44}/\textbf{0.7741}
\\
9 & 23.06/0.6636 & 23.27/0.6754 & 23.31/0.6784 & 23.26/0.6757 & 23.36/0.6822 & 23.30/0.6789 & \textbf{23.55}/\textbf{0.6903}
\\
10 & 37.39/0.8860 & 37.48/0.8877 & 37.51/0.8884 & 37.50/0.8883 & 37.52/0.8885 & 37.50/0.8882 & \textbf{37.72}/\textbf{0.8950}
\\
11 & 34.65/0.8620 & 34.90/0.8673 & 34.92/0.8676 & 34.90/0.8670 & 34.97/0.8688 & 34.93/0.8679 & \textbf{35.05}/\textbf{0.8721}
\\
12 & 27.50/0.6394 & 27.61/0.6457 & 27.62/0.6482 & 27.60/0.6452 & 27.66/0.6519 & 27.64/0.6495 & \textbf{27.72}/\textbf{0.6565}
\\
13 & 26.16/0.7576 & 26.53/0.7741 & 26.55/0.7762 & 26.50/0.7731 & 26.65/0.7807 & 26.56/0.7763 & \textbf{26.81}/\textbf{0.7869}
\\
14 & 33.27/0.7518 & 33.35/0.7539 & 33.38/0.7551 & 33.33/0.7546 & 33.35/0.7555 & 33.36/0.7545 & \textbf{33.45}/\textbf{0.7602}
\\
15 & 25.48/0.6900 & 25.80/0.7014 & 25.84/0.7043 & 25.80/0.7025 & 25.90/0.7069 & 25.83/0.7055 & \textbf{26.00}/\textbf{0.7124}
\\
16 & 27.97/0.7213 & 28.06/0.7250 & 28.07/0.7255 & 28.06/0.7251 & 28.10/0.7272 & 28.08/0.7258 & \textbf{28.19}/\textbf{0.7320}
\\
17 & 28.74/0.7631 & 28.89/0.7689 & 28.91/0.7699 & 28.88/0.7689 & 28.95/0.7719 & 28.91/0.7702 & \textbf{29.03}/\textbf{0.7762}
\\
18 & 23.36/0.7927 & 24.24/0.8179 & 24.31/0.8203 & 24.20/0.8171 & 24.62/0.8280 & 24.36/0.8209 & \textbf{24.94}/\textbf{0.8368}
\\
19 & 36.85/0.8920 & 37.28/0.8992 & 37.30/0.8995 & 37.23/0.8987 & 37.43/0.9015 & 37.25/0.8989 & \textbf{37.59}/\textbf{0.9056}
\\
20 & 38.02/0.9129 & 38.18/0.9146 & 38.17/0.9147 & 38.15/0.9147 & 38.26/0.9156 & 38.17/0.9149 & \textbf{38.33}/\textbf{0.9182}
\\
21 & 26.89/0.8433 & 27.38/0.8545 & 27.39/0.8554 & 27.33/0.8541 & 27.48/0.8579 & 27.39/0.8562 & \textbf{27.68}/\textbf{0.8636}
\\
22 & 27.68/0.8091 & 28.19/0.8263 & 28.21/0.8265 & 28.13/0.8239 & 28.31/0.8300 & 28.20/0.8256 & \textbf{28.48}/\textbf{0.8351}
\\
23 & 26.26/0.7492 & 26.51/0.7588 & 26.55/0.7606 & 26.50/0.7589 & 26.64/0.7645 & 26.54/0.7611 & \textbf{26.72}/\textbf{0.7682}
\\
24 & 30.09/0.7624 & 30.22/0.7659 & 30.23/0.7670 & 30.22/0.7665 & 30.26/0.7680 & 30.23/0.7666 & \textbf{30.33}/\textbf{0.7718}
\\
25 & 25.07/0.7363 & 25.35/0.7471 & 25.41/0.7499 & 25.36/0.7475 & 25.47/0.7525 & 25.41/0.7504 & \textbf{25.62}/\textbf{0.7585}
\\
26 & 24.39/0.6501 & 24.61/0.6580 & 24.64/0.6604 & 24.61/0.6588 & 24.72/0.6637 & 24.66/0.6616 & \textbf{24.80}/\textbf{0.6676}
\\
27 & 29.42/0.8154 & 29.74/0.8258 & 29.81/0.8276 & 29.72/0.8255 & 29.86/0.8306 & 29.80/0.8272 & \textbf{30.03}/\textbf{0.8354}
\\
28 & 33.46/0.8813 & 33.99/0.8921 & 34.09/0.8930 & 33.94/0.8905 & 34.20/0.8944 & 33.99/0.8898 & \textbf{34.53}/\textbf{0.9007}
\\
29 & 24.60/0.7345 & 25.04/0.7488 & 25.08/0.7505 & 25.02/0.7481 & 25.16/0.7544 & 25.06/0.7507 & \textbf{25.34}/\textbf{0.7598}
\\
30 & 27.60/0.7737 & 28.12/0.7944 & 28.17/0.7966 & 28.10/0.7934 & 28.26/0.8005 & 28.17/0.7964 & \textbf{28.39}/\textbf{0.8054}
\\
\midrule
Average & 29.26/0.7850 & 29.57/0.7946 & 29.60/0.7960 & 29.55/0.7944 & 29.68/0.7987 & 29.60/0.7961 & \textbf{29.82}/\textbf{0.8018}
\\
\bottomrule
\end{tabular}
}
\end{table*}

\begin{figure*}[!htpb]
\begin{minipage}[b]{\linewidth}
\centering
% \resizebox{宽度}{高度}{对象}
\resizebox{0.95\linewidth}{!}{
\includegraphics[width=\linewidth]{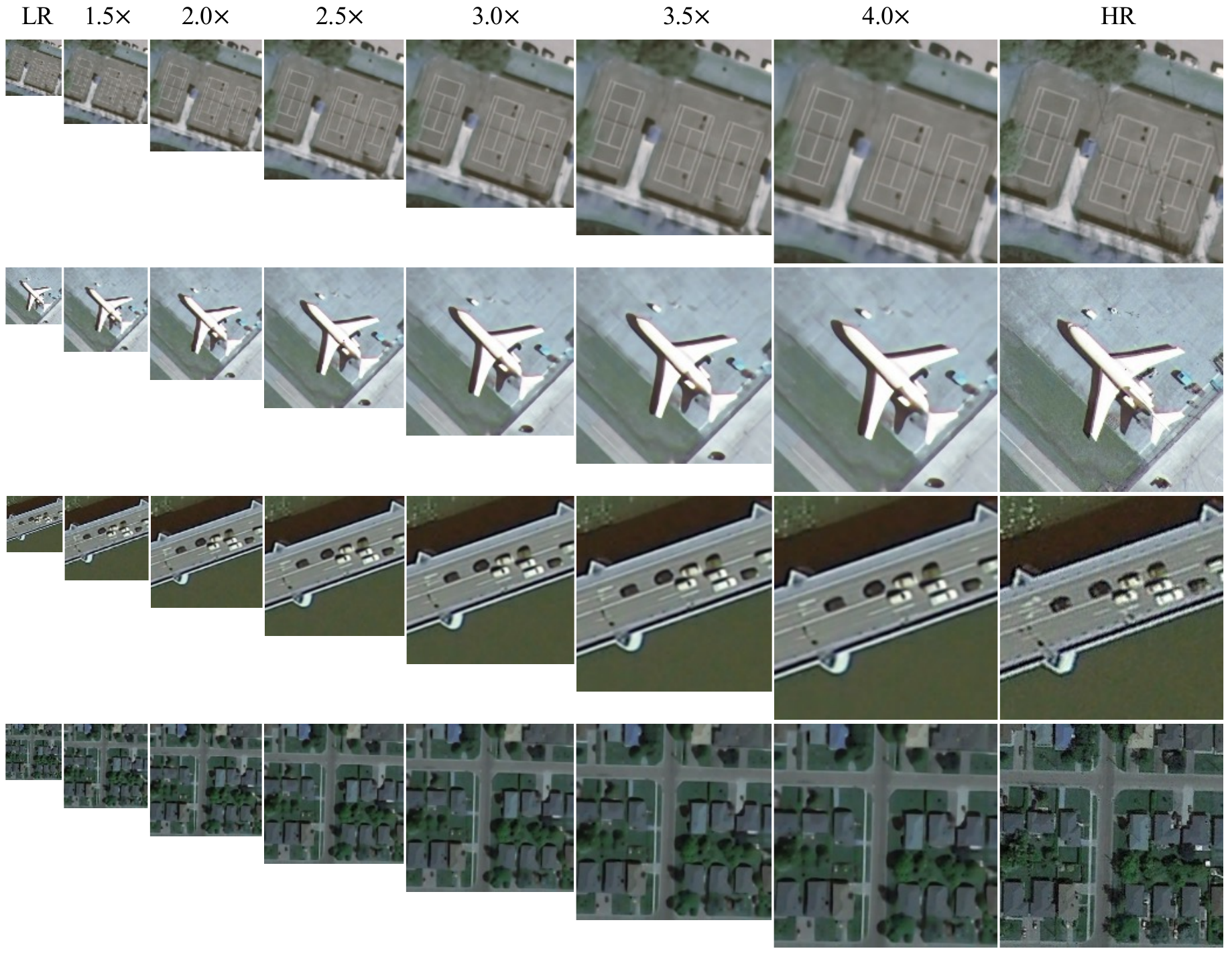}
}
\caption{The visual comparisons of some image examples upsampling with different scale factors by FunSR-RDN. 
The LR image is downsampled from the HR reference image with a scale ratio of $1/4$. 
The first two rows are from the UCMerced test set (``tenniscourt99" and ``airplane35"), while the last two are from the AID test set (``bridge\_28" and ``denseresidential\_20").}
\label{fig:continuous_results}
\end{minipage}
\vfill
\vspace{6pt}
\begin{minipage}[b]{\linewidth}
\centering
% \resizebox{宽度}{高度}{对象}
\resizebox{0.95\linewidth}{!}{
\includegraphics[width=\linewidth]{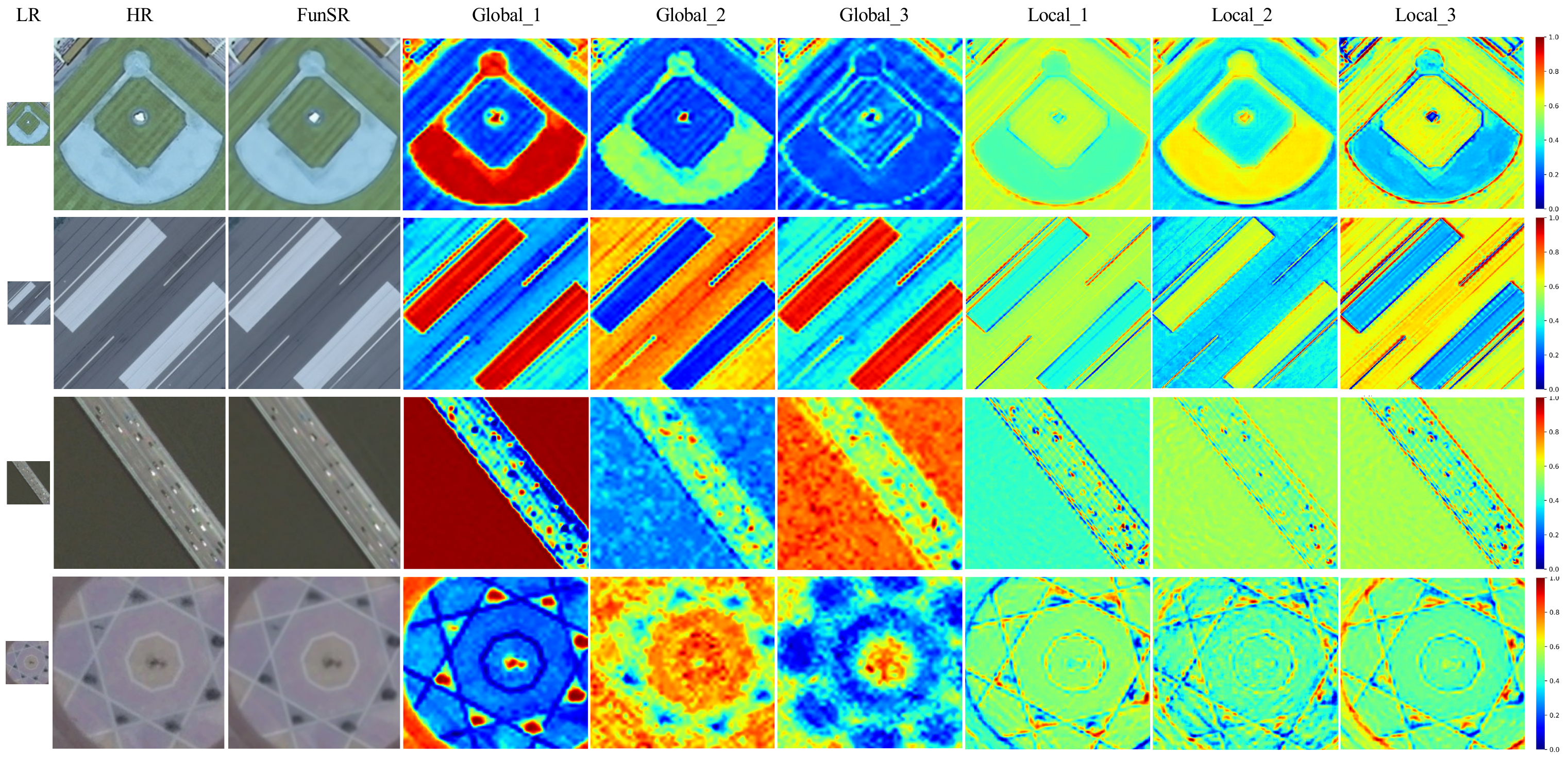}
}
\caption{Some visual examples of the feature maps generated from the global and local parsers with an upscale factor of $\times 4$. The first two rows are scenes (``baseballdiamond31" and ``runway76") from the UCMerced dataset and the last two rows (``bridge\_353" and
``square\_83") are from the AID dataset.}
\label{fig:local_global}
\end{minipage}
\end{figure*}

\begin{figure*}[!htpb]
\begin{minipage}[b]{\linewidth}
\centering
% \resizebox{宽度}{高度}{对象}
\resizebox{0.93\linewidth}{!}{
\includegraphics[width=\linewidth]{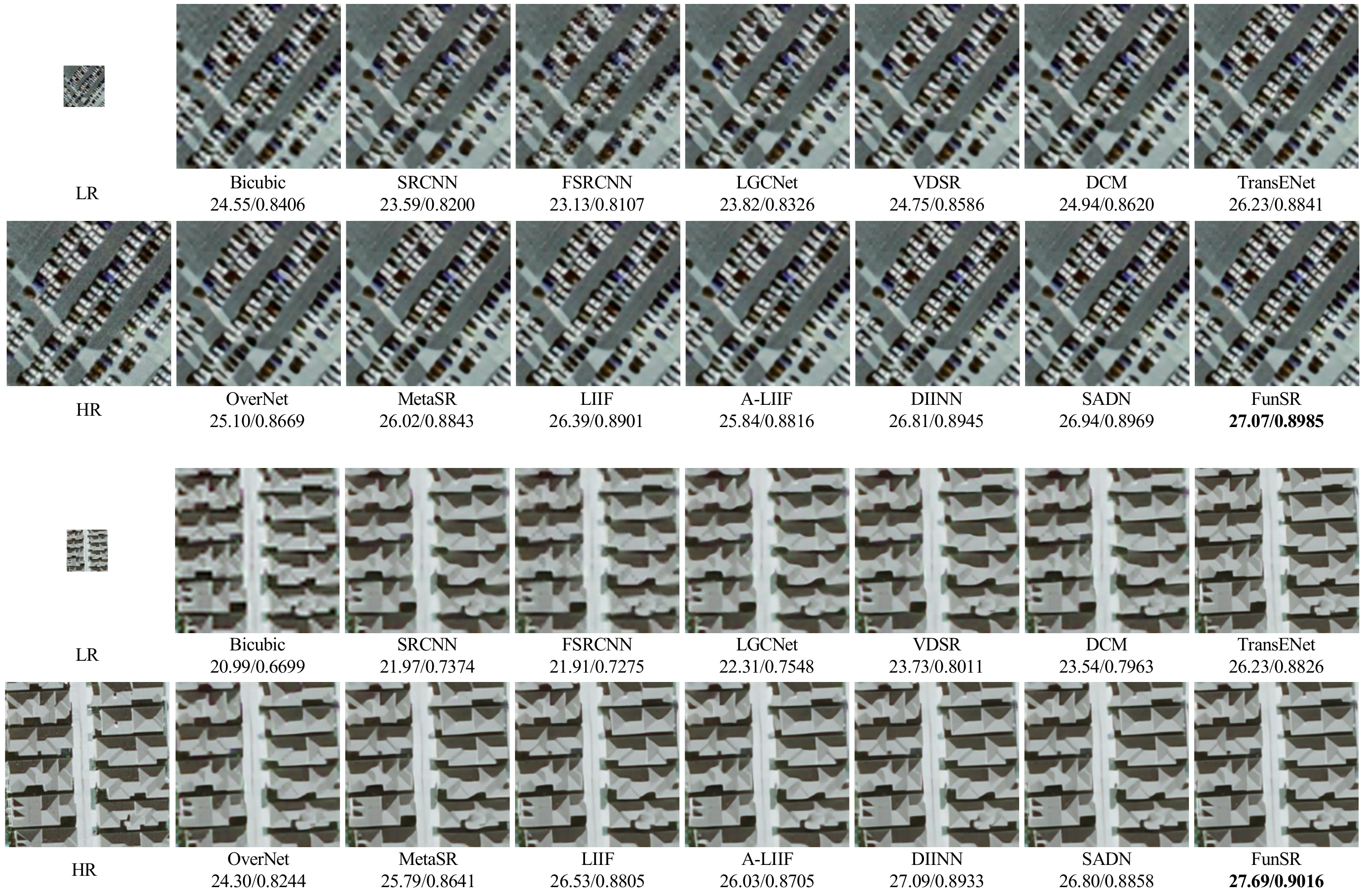}
}
\caption{Comparisons on the UCMerced test set with different methods under $\times 4$ factor. Image crops are from ``parkinglot17" and ``denseresidential58" respectively. Zoom in for better visualization.}
\label{fig:uc_compare_vis}
\end{minipage}
\vfill
\vspace{6pt}
\begin{minipage}[b]{\linewidth}
\centering
% \resizebox{宽度}{高度}{对象}
\resizebox{0.93\linewidth}{!}{
\includegraphics[width=\linewidth]{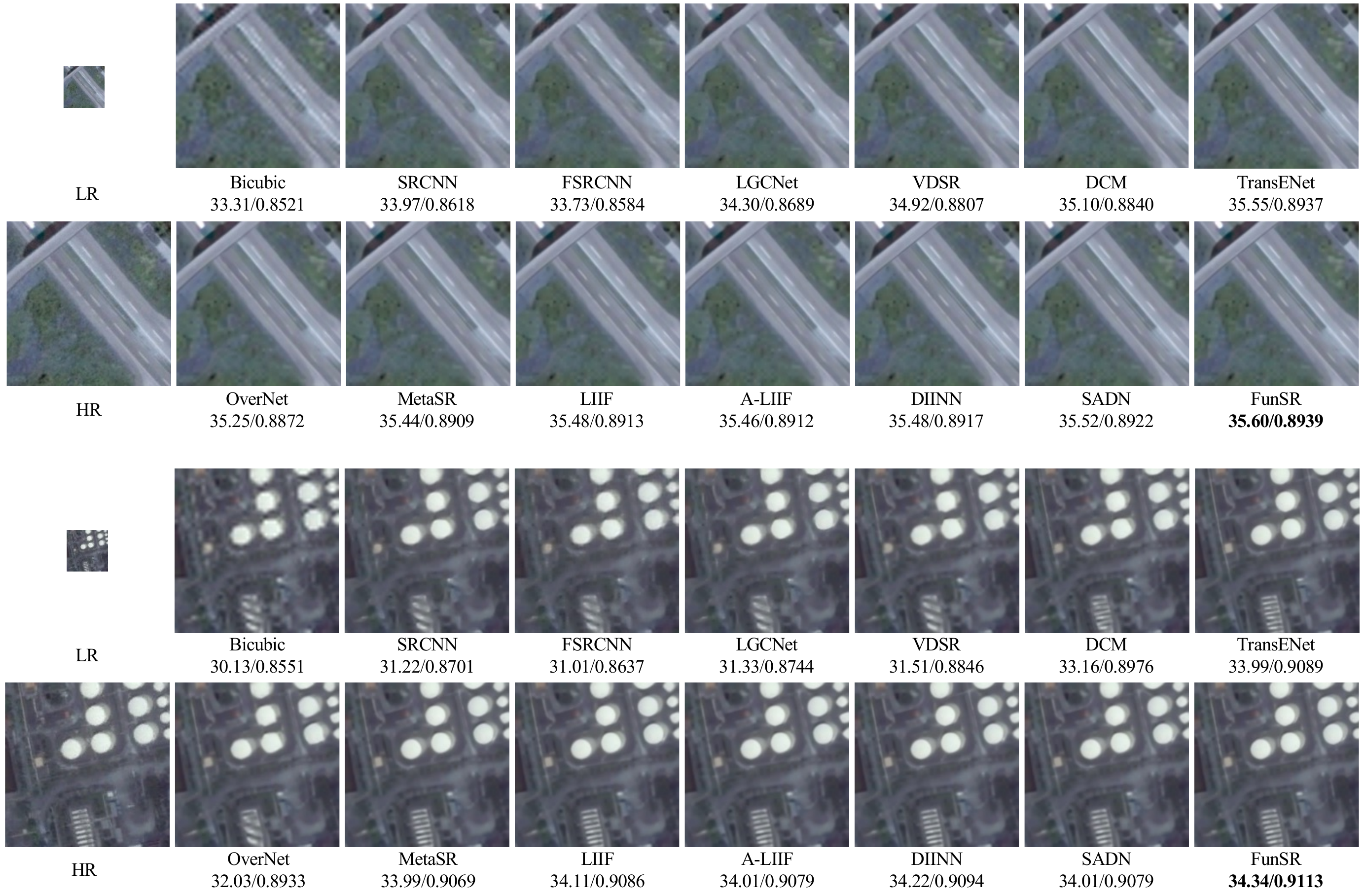}
}
\caption{Comparisons on the AID test set with different methods under $\times 4$ factor. Image crops are from ``viaduct\_271" and ``storagetanks\_336" respectively. Zoom in for better visualization.}
\label{fig:aid_compare_vis}
\end{minipage}
\end{figure*}

\subsubsection{Training Details}

During the training phase, $48 \times 48$ patches are randomly cropped from LR remote sensing images, and reference patches are cropped from their corresponding HR ones.
Meanwhile, we use random rotation ($90^\circ$, $180^\circ$, and $270^\circ$) and horizontal flipping to augment the training samples. 
The coordinate and image RGB values are all normalized to be between -1 and 1. 
The scale factors in the training phase are uniformly distributed from 1 to 4. 
During the test phase, the LR test images are cropped into a set of $48 \times 48$ patches and then fed into the model without any bells and whistles.

For optimization, we use AdamW optimizer with an initial learning rate of $1e-4$ to train our model. The mini-batch size is set to 8. The total training epochs are 4000 for the UCMecred dataset and 2000 for the AID dataset. We conduct a Cosine Annealing scheduler \cite{loshchilov2016sgdr} to decay the learning rate. The proposed method is implemented by PyTorch, and all experiments are run on an NVIDIA A100 Tensor Core GPU.

\subsection{Comparison with the State-of-the-Art}

In this section, we compare the proposed method with some other state-of-the-art SR methods, including the classic bicubic interpolation, fixed magnification SR methods (\textit{e.g.}, SRCNN \cite{dong2015image}, FSRCNN \cite{dong2016accelerating}, LGCNet \cite{lei2017super}, VDSR \cite{kim2016accurate}, DCM \cite{haut2019remote}, TransENet \cite{lei2021transformer}), and continuous magnification SR methods (\textit{e.g.}, MetaSR \cite{hu2019meta}, LIIF \cite{chen2021learning}, A-LIIF \cite{li2022adaptive}, DIINN \cite{nguyen2023single}, SADN \cite{wu2021scale}, OverNet \cite{behjati2021overnet}, ArbRCAN \cite{wang2021learning}). 
In terms of continuous magnification image SR, we have applied various image encoders (EDSR \cite{lim2017enhanced}, RCAN \cite{zhang2018image}, and RDN \cite{zhang2018residual}) to verify the robustness of the proposed method.
We can run out-of-distribution evaluations (training under $\times 4$, evaluating beyond $\times 4$) because of the coordinate-based architecture.
All the methods are implemented according to the official publications with Pytorch.

\subsubsection{Quantitative Results on the UCMerced Dataset}

The results of FunSR versus other comparison methods on the UCMerced Dataset are shown in Tab. \ref{tab:uc_sota_allx}, with the best performance shown by a bold number.
We just show the upscale factor of $\times 2.0$, $\times 2.5$, $\times 3.0$, $\times 3.5$, $\times 4.0$, $\times 6.0$, $\times 8.0$, and $\times 10.0$ for simplicity. 
FunSR nearly achieves the highest performance in terms of PSNR and SSIM across all backbones and upscale factors.
Specifically, 
FunSR outperforms the state-of-the-art fixed magnification transformer-based SR method TransENet (26.98/0.7755) by 27.11/0.7781, 27.24/0.7799, and 27.29/0.7798 on PSNR and SSIM under $\times 4$ magnification utilizing EDSR, RCAN, and RDN image encoders, respectively.
FunSR has also shown comparable performance with continuous image SR algorithms over different backbones for in-distribution and out-of-distribution training magnifications.

The comprehensive results of various methods for all 21 scene classes\footnote{All the 21 classes of the UCMerced dataset: 1 agricultural, 2 airplane, 3 baseballdiamond, 4 beach, 5 buildings, 6 chaparral, 7 denseresidential, 8 forest, 9 freeway, 10 golfcourse, 11 harbor, 12 intersection, 13 mediumresidential, 14 mobilehomepark, 15 overpass, 16 parkinglot, 17 river, 18 runway, 19 sparseresidential, 20 storagetanks, and 21 tenniscourt.} of the UCMeced dataset are available in Tab. \ref{tab:uc_sota_classes} at $\times 4$ magnification.
We can see that FunSR achieves the best PSNR/SSIM values in most scene classes, whereas TransENet achieves comparable performance in the remaining four, namely agricultural, beach, chaparral, and golfcourse.
When compared to the TransENet, FunSR is more effective in situations with high-frequency characteristics and repetitive patterns, such as buildings, dense residences, storage tanks, and tennis courts.

\begin{figure}[!tpb]
\centering
% \resizebox{宽度}{高度}{对象}
\resizebox{\linewidth}{!}{
\includegraphics[width=\linewidth]{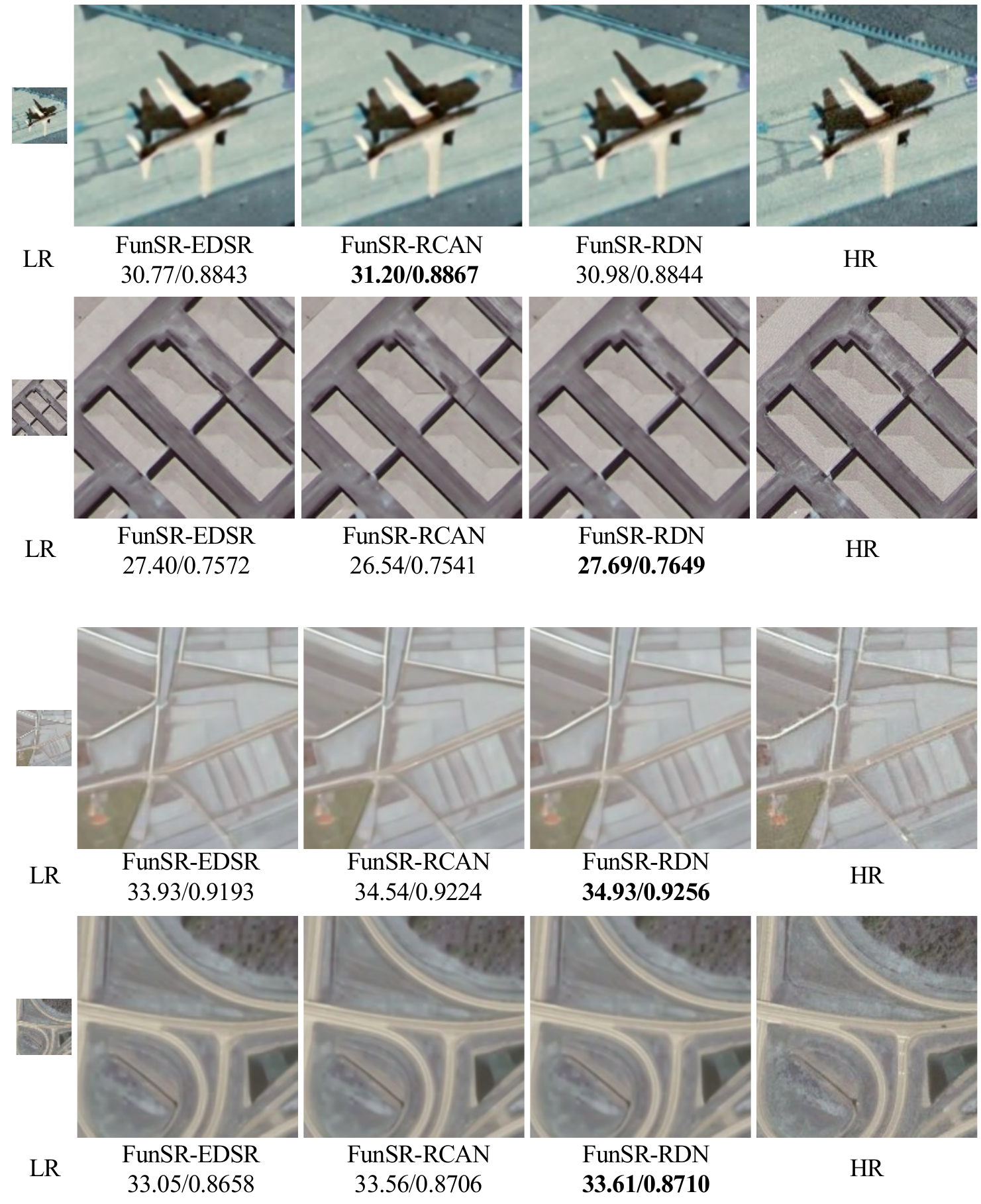}
}
\caption{Some SR results on the UCMerced dataset (top two rows) and the AID dataset (bottom two rows) with different image encoders under the $\times 4$ factor. Image crops are from ``airplane15", ``buildings25", ``farmland\_112", and ``viaduct\_108" respectively. Zoom in for better visualization.}
\label{fig:vis_funsr}
\end{figure}

\subsubsection{Quantitative Results on AID Dataset}

We conduct comparison experiments on the AID dataset to further validate FunSR's effectiveness. 
Unlike the UCMerced dataset, this one is larger in size and has more scene categories, totaling 30.
Tab. \ref{tab:aid_sota_allx} displays the overall results of various methods on this dataset.
It can be seen that, when compared to other approaches, FunSR produces the best results on the majority of magnifications presented across different image encoders.

In addition, Tab. \ref{tab:aid_sota_classes} lists the detailed performance on the 30 classes\footnote{All the 30 classes of the AID dataset: 1 Airport, 2 BareLand, 3 BaseballField, 4 Beach, 5 Bridge, 6 Center, 7 Church, 8 Commercial, 9 DenseResidential, 10 Desert, 11 Farmland, 12 Forest, 13 Industrial, 14 Meadow, 15 MediumResidential, 16 Mountain, 17 Park, 18 Parking, 19 Playground, 20 Pond, 21 Port, 22 RailwayStation, 23 Resort, 24 River, 25 School, 26 SparseResidential, 27 Square, 28 Stadium, 29 StorageTanks, and 30 Viaduct.} with a magnification of $\times 4$. It demonstrates that FunSR delivers the best results on all the surface target scenes. According to Tab. \ref{tab:aid_sota_allx} and Tab. \ref{tab:aid_sota_classes}, it implies that FunSR can produce better results as the quantity of the dataset and the complexity of the data expand.

\subsubsection{Qualitative Visual Comparisons}

In addition to quantitative comparison, 
we have presented some visual examples from FunSR-RDN on continuous magnification factors in Fig. \ref{fig:continuous_results}.
It can be observed that the proposed FunSR can be used to generate considerable visual effects in image SR at continuous magnification.
For a better visual comparison with other methods, we here provide a qualitative comparison of the recovered results under the upscale factor of $\times 4$. 
Fig. \ref{fig:uc_compare_vis} displays some super-resolved examples of the UCMerced dataset including ``parking lot" and ``dense residential” scenes, while Fig.~\ref{fig:aid_compare_vis} presents some of the AID dataset, including ``viaduct" and ``storage tanks” scenes. 
Overall, when compared to other approaches, the proposed FunSR produces better results with crisper edges and contours that are also closer to the HR references.

\begin{table}[!tpb] 
\begin{minipage}[b]{\linewidth}
\centering
\caption{Ablation of various components in FunSR-EDSR. Mean PSNR (dB) and SSIM on the UC test dataset with continuous upscale factors are provided.
}
\label{tab:uc_ablation_components}
\resizebox{\linewidth}{!}{
\begin{tabular}{*{3}{c} | *{5}{c}}
\toprule
$\Phi_\text{pyramid}$ & $\Phi_\text{interactor}$ & $\Phi_\text{global}$ & $\times$2.0 & $\times$2.5& $\times$3.0& $\times$3.5& $\times$4.0
\\
\midrule
& &  &
34.01/0.9251
&30.83/0.8759
&28.61/0.8247
&27.50/0.7946
&26.46/0.7574
\\
\checkmark &  &  
&34.36/0.9279
&31.15/0.8823
&28.97/0.8365
&27.86/0.8054
&26.86/0.7733
\\
\checkmark & \checkmark & 
&34.49/0.9306
&31.27/0.8847
&29.10/0.8389
&27.98/0.8077
&26.95/0.7751
\\
\checkmark & \checkmark & \checkmark &\textbf{34.61}/\textbf{0.9318}
&\textbf{31.40}/\textbf{0.8860}
&\textbf{29.19}/\textbf{0.8391}
&\textbf{28.10}/\textbf{0.8095}
&\textbf{27.11}/\textbf{0.7781}
\\
\bottomrule
\end{tabular}
}
\end{minipage}
\vfill
\vspace{5pt}
\begin{minipage}[b]{\linewidth}
\centering
\caption{Ablation of the positional encoding (PE) and the learnable global token (LT) in the functional interactor.}
\label{tab:uc_ablation_interactor}
\resizebox{\linewidth}{!}{
\begin{tabular}{*{2}{c} | *{5}{c}}
\toprule
PE & LT & $\times$2.0 & $\times$2.5& $\times$3.0& $\times$3.5& $\times$4.0
\\
\midrule
\checkmark & 
&32.07/0.9024
&28.91/0.8366
&26.77/0.7731
&25.64/0.7271
&24.71/0.6855
\\
& \checkmark
&34.28/0.9283
&31.06/0.8825
&28.98/0.8358
&27.75/0.8047
&26.75/0.7731
\\
\checkmark & \checkmark &\textbf{34.61}/\textbf{0.9318}
&\textbf{31.40}/\textbf{0.8860}
&\textbf{29.19}/\textbf{0.8391}
&\textbf{28.10}/\textbf{0.8095}
&\textbf{27.11}/\textbf{0.7781}
\\
\bottomrule
\end{tabular}
}
\end{minipage}
\vfill
\vspace{5pt}
\begin{minipage}[b]{\linewidth}
\centering
\caption{Ablation of different input attributes (GC: global coordinate, LC: local coordinate, SR: scale ratio, RGB: interpolated RGB value) in the functional parser.}
\label{tab:uc_ablation_parser}
\resizebox{\linewidth}{!}{
\begin{tabular}{*{4}{c} | *{5}{c}}
\toprule
GC & LC & SR & RGB & $\times$2.0 & $\times$2.5& $\times$3.0& $\times$3.5& $\times$4.0
\\
\midrule
\checkmark &  &  & 
&33.97/0.9245
&30.63/0.8709
&28.52/0.8184
&27.37/0.7877
&26.43/0.7497
\\
\checkmark & \checkmark &  &  
&34.39/0.9292
&31.03/0.8782
&28.82/0.8271
&27.71/0.7960
&26.67/0.7601
\\
\checkmark & \checkmark & \checkmark &
&34.54/0.9317
&31.17/0.8816
&29.01/0.8308
&27.87/0.7990
&26.86/0.7657
\\
\checkmark & \checkmark & \checkmark & \checkmark &\textbf{34.61}/\textbf{0.9318}
&\textbf{31.40}/\textbf{0.8860}
&\textbf{29.19}/\textbf{0.8391}
&\textbf{28.10}/\textbf{0.8095}
&\textbf{27.11}/\textbf{0.7781}
\\
\bottomrule
\end{tabular}
}
\end{minipage}
\end{table}

\subsection{Ablation Study}

In this section, we run a series of experiments on the UCMereced dataset to explore the significance of each component in our method, with all models trained with the same settings of the EDSR image encoder unless otherwise specified.

\subsubsection{Effects of Different Components in FunSR}

We conducted relevant ablation experiments with FunSR-EDSR on the UCMerced dataset to validate the effectiveness of components in the proposed FunSR.
As seen in Tab. \ref{tab:uc_ablation_components}, performance tends to grow monotonically
with the increased component terms, from 26.46/0.7574 to
27.11/0.7781 on PSNR/SSIM metric under $\times 4$ upscale factor, suggesting that all the designed components count. 
To intuitively depict the role of the local and the global parsers, Fig. \ref{fig:local_global} shows the final feature map ($2 \times 3$ feature maps) generated by the two parsers. It can be seen that the global parser prioritizes the restoration of the overall scene, whereas the local parser pays more attention to details, such as contours, textures, etc.

\subsubsection{Effects of Different Image Encoders in the Functional Representor}

We have tried different image encoders on the LR images in the proposed FunSR to learn shallow functional representations. 
Tab. \ref{tab:uc_sota_allx} and Tab. \ref{tab:aid_sota_allx} display the performance values of different encoders on the two datasets respectively. 
Experiments show that FunSR can outperform other methods across different image encoders.
Fig. \ref{fig:vis_funsr} provides a straightforward visualization for comparing SR results among different image encoders.

\subsubsection{Effects of Positional Encoding and Learnable Global Token in the Functional Interactor}

The functional interactor is designed to allow each pixel-wise function to interact with functions in other locations, hence improving global semantic coherence.
To actualize the interaction process, we use a Transformer-based design.
We conducted ablation studies on the positional encoding and the learnable global token to validate the necessity for its internal detailed designs.
For the design without a learnable global token, we use a global pooling operation on the feature map generated by the interactor to form the global function parameters ($\theta_{\text{global}}$). 
Tab. \ref{tab:uc_ablation_interactor} shows the ablation results on the UCMerced dataset using FunSR-EDSR. It indicates that FunSR performs significantly worse without the global learnable token and marginally worse without the positional encoding.

\begin{figure}[!tpb]
\begin{minipage}[b]{\linewidth}
\centering
% \resizebox{宽度}{高度}{对象}
\resizebox{0.9\linewidth}{!}{
\includegraphics[width=\linewidth]{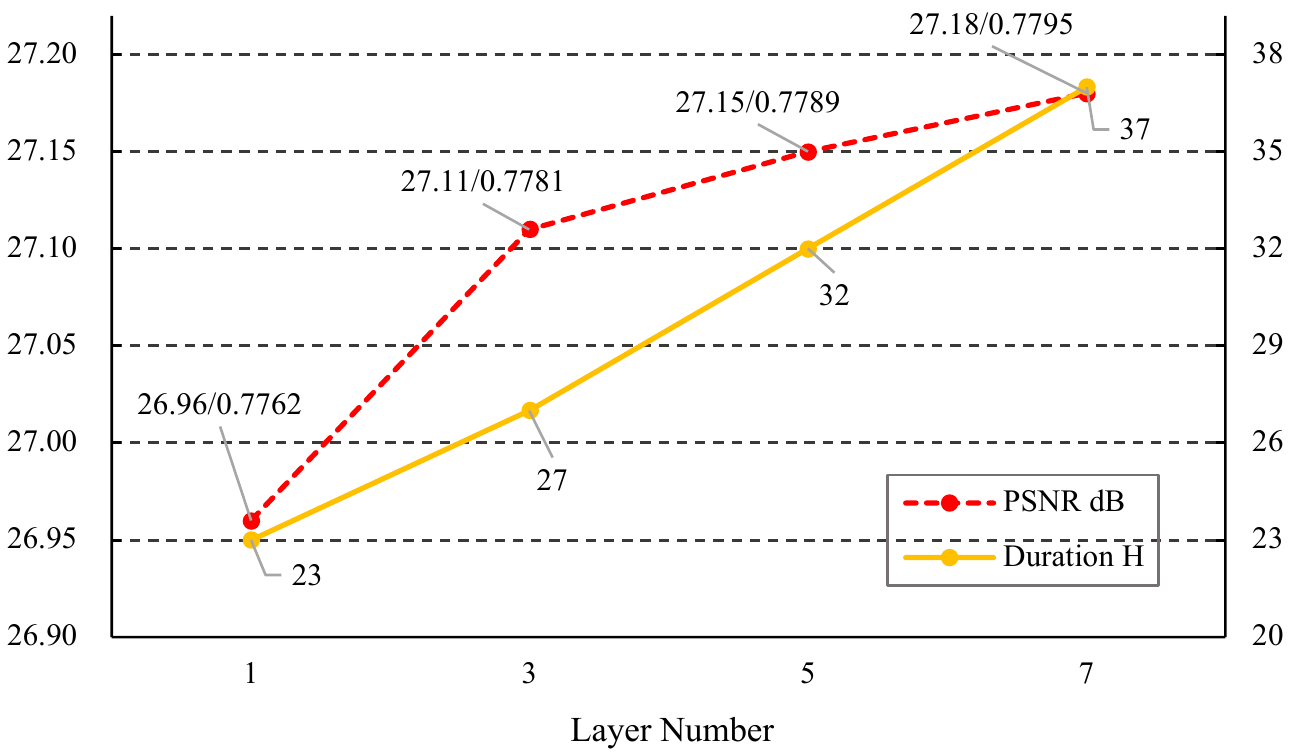}
}
\caption{PSNR/SSIM and convergence duration time over different layer numbers of the interactor.}
\label{fig:number_of_interactor}
\end{minipage}
\vfill
\vspace{7pt}
\begin{minipage}[b]{\linewidth}
\centering
% \resizebox{宽度}{高度}{对象}
\resizebox{0.9\linewidth}{!}{
\includegraphics[width=\linewidth]{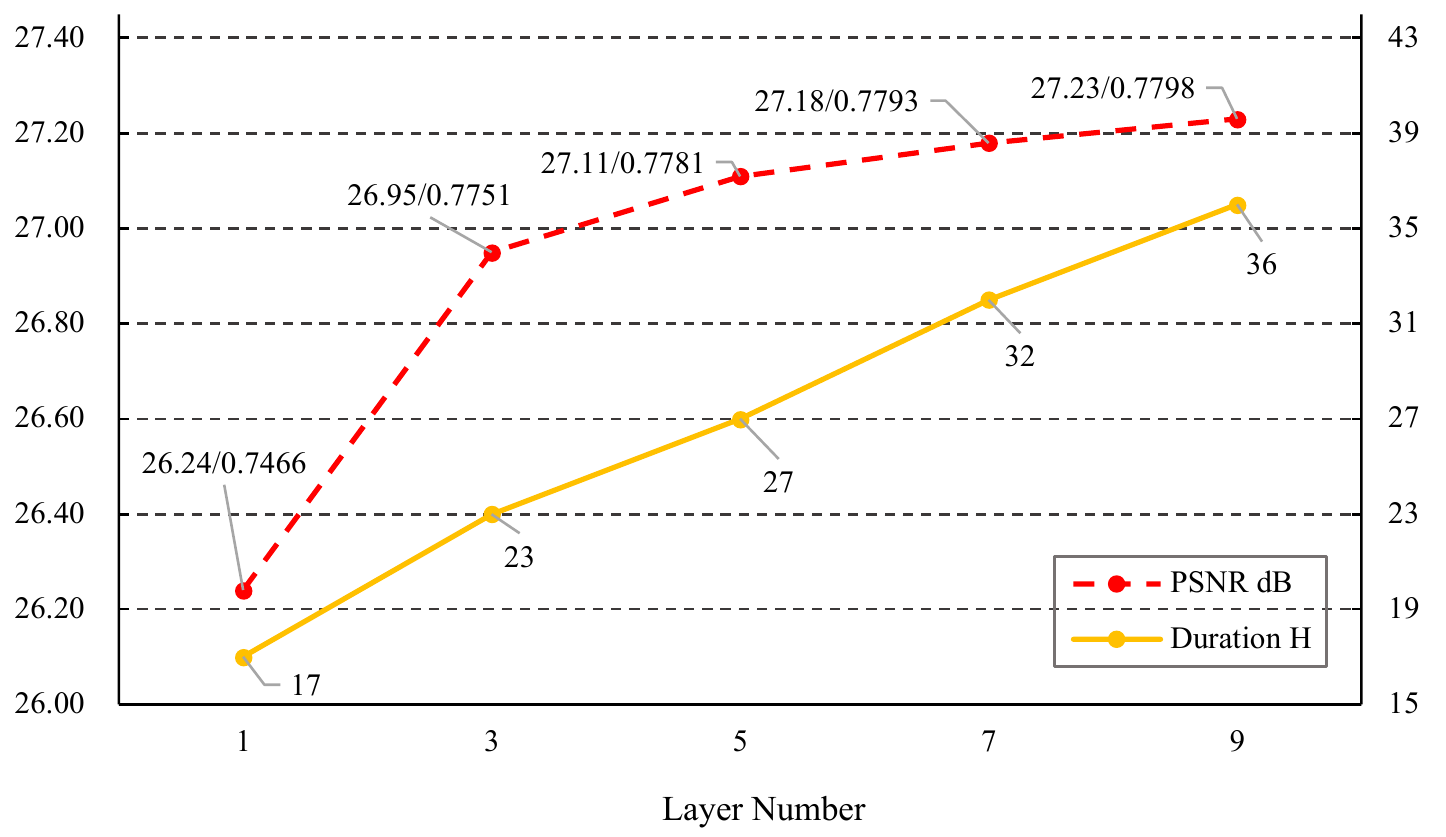}
}
\caption{PSNR/SSIM and convergence duration time over different layer numbers of the parser.}
\label{fig:number_of_parser}
\end{minipage}
\end{figure}

% \begin{table}[!tpb] 
% \centering
% \caption{Effect of different interpolation methods for getting local parser’s parameters.}
% \label{tab:uc_ablation_interpolation}
% \resizebox{\linewidth}{!}{
% \begin{tabular}{*{3}{c} | *{5}{c}}
% \toprule
%  Nearest & Bilinear & Bicubic & $\times$2.0 & $\times$2.5& $\times$3.0& $\times$3.5& $\times$4.0
% \\
% \midrule
% \checkmark &  &
% &34.61/0.9318
% &31.40/0.8860
% &29.19/0.8391
% &28.10/0.8095
% &27.11/0.7781
% \\
%  & \checkmark &
% &34.62/0.9317
% &31.43/0.8862
% &29.20/0.8390
% &28.12/0.8096
% &27.12/0.7784
% \\
%  &  & \checkmark
% &34.67/0.9323
% &31.46/0.8868
% &29.22/0.8394
% &28.14/0.8099
% &27.15/0.7787 
% \\
% \bottomrule
% \end{tabular}
% }
% \end{table}

\subsubsection{Effects of Different Input Attributes in the Functional Parser}

The functional parser queries the pixel-wise functions based on the corresponding coordinates and parses the RGB pixel value of the position by inputting the global coordinates and certain extra properties, such as local coordinates, scale factor, and interpolated RGB value.
Since it doesn't make sense to feed the local coordinates into the global parser, we only consider global coordinates in the global parser. 
Tab. \ref{tab:uc_ablation_interactor} indicates that as the input information increases, the SR performance also increases monotonically, 
\textit{e.g.}, from 26.43/0.7497 to 27.11/0.7781 on PNSR/SSIM under the magnification of $\times 4$.

% \subsubsection{Different Interpolation Methods for Getting Local Parser's Parameters}

% Because the function parameter map ($\theta_{\text{local}}$) received by the functional interactor is low-resolution, we have to employ an interpolation operation to obtain the parameters of the local parser, coping with the arbitrary resolution output. Tab. \ref{tab:uc_ablation_interpolation} demonstrates that different interpolation methods have little impact on the SR performance. We apply the nearest interpolation for efficiency.

\subsubsection{Layer Numbers of the Interactor and Parser}

The layer numbers of the interactor and parser can have an effect on our method's performance.
Therefore, we perform a series of experiments on this topic. 
In addition to the effectiveness metrics (PNSR and SSIM), we also pay attention to the efficiency, because the Transformer in the interactor and the MLP in the parser are resource consumption.
Here we take the convergence duration time (H, Hour) during training to measure the efficiency. We conduct the experiments on a single NVIDIA A100 Tensor Core GPU with a batch size of 8. 
Fig. \ref{fig:number_of_interactor} and Fig. \ref{fig:number_of_parser} give a clear understanding of the PSNR/SSIM and convergence duration time over different layer numbers of the interactor and parser respectively. 
To balance efficiency and effectiveness, we finally choose a 3-layer interactor and a 5-layer parser to form the proposed FunSR.

\section{Conclusion}

In this paper, we propose FunSR, a novel SR framework for remote sensing images. FunSR aims at learning continuous representations for remote sensing images based on context interaction in implicit function space. It consists of three main parts: a functional representor, a functional interactor, and a functional parser. 
The representor first converts the LR image to a multi-scale continuous function representation, then the interactor allows each pixel-wise function to interact with functions at other locations and levels, and finally, the parser parses the discrete coordinates of the HR image to corresponding RGB values. 
The effectiveness of each component has been verified through ablation studies. Meanwhile, experimental results on two public datasets reveal that our method outperforms other state-of-the-art fixed magnification and continuous magnification methods in terms of super-resolved results.

% Can use something like this to put references on a page
% by themselves when using endfloat and the captionsoff option.
\ifCLASSOPTIONcaptionsoff
  \newpage
\fi

% trigger a \newpage just before the given reference
% number - used to balance the columns on the last page
% adjust value as needed - may need to be readjusted if
% the document is modified later
%\IEEEtriggeratref{8}
% The "triggered" command can be changed if desired:
%\IEEEtriggercmd{\enlargethispage{-5in}}

% references section

% can use a bibliography generated by BibTeX as a .bbl file
% BibTeX documentation can be easily obtained at:
% http://mirror.ctan.org/biblio/bibtex/contrib/doc/
% The IEEEtran BibTeX style support page is at:
% http://www.michaelshell.org/tex/ieeetran/bibtex/
\bibliographystyle{IEEEtran}
% argument is your BibTeX string definitions and bibliography database(s)
\bibliography{IEEEabrv,myreferences}
\end{document}